\documentclass[english,10pt]{article}
\usepackage[papersize={19cm,25cm},body={15cm,21cm},centering]{geometry}
\usepackage[T1]{fontenc}
\usepackage[latin9]{inputenc}
\usepackage{amsmath}
\usepackage{amssymb}
\usepackage{url}
\usepackage{babel}
 \usepackage{graphicx}
\usepackage{epsfig,subfigure,color}
\usepackage{graphics}
 \usepackage{graphicx}
\usepackage{amsmath,amsfonts,amsthm,latexsym}
\usepackage{mathrsfs,amsbsy}
\usepackage{multirow}
\usepackage{epstopdf}
\usepackage{algorithm}
\usepackage{algorithmic}
\usepackage{cite}

\newtheorem{property}{Property}
\newtheorem{example}{Example}
\newtheorem{remark}{Remark}

\theoremstyle{definition}
\newtheorem{defn}{Definition}

\DeclareMathOperator{\tr}{trace}

\begin{document}

\title{ Co-Separable Nonnegative Matrix Factorization}
\date{}
\author{
Junjun Pan \qquad  Michael K. Ng\thanks{Department of Mathematics,
The University of Hong Kong.
Emails: junjpan@hku.hk, mng@maths.hku.hk. M. Ng's research is supported
in part by HKRGC GRF 12300218, 12300519, 17201020 and 17300021.
} 
}

\maketitle

\begin{abstract}

Nonnegative matrix factorization (NMF) is a popular model in the field of pattern
recognition. It aims to find a low rank approximation for nonnegative data $M$ by a product of two nonnegative matrices $W$ and $H$. In general, NMF is NP-hard to solve while it can be solved efficiently under separability assumption, which requires the columns of factor matrix are equal to columns of the input matrix. In this paper, we generalize separability assumption based on 3-factor NMF $M=P_1SP_2$, and require that $S$ is a sub-matrix of the input matrix. We refer to this NMF 
as a Co-Separable NMF (CoS-NMF). We discuss  
some mathematics properties of CoS-NMF, and present the relationships with other 
related matrix factorizations such as CUR decomposition, generalized separable NMF(GS-NMF), and  
bi-orthogonal tri-factorization (BiOR-NM3F). An optimization model for CoS-NMF 
is proposed and alternated fast gradient method is employed to solve the model. 
Numerical experiments 
on synthetic datasets, document datasets and facial databases are conducted 
to verify the effectiveness of our CoS-NMF model. 
Compared to state-of-the-art methods, CoS-NMF model performs very well 
in co-clustering task, and preserves a good approximation to the input data matrix 
as well.
\end{abstract}


\section{Introduction}
Matrix methods lie at the root of most methods of machine learning and data analysis. Among all matrix methods, nonnegative matrix factorization (NMF) is an important one. It can automatically extracts sparse and meaningful features from a set of nonnegative data vectors and has become a popular tool in data mining society. Given a nonnegative matrix $M\in \mathbb{R}^{m\times n}_+$ and an integer factorization rank $r$, NMF is the problem of computing $W\in \mathbb{R}^{m\times r}_+$ and $H\in \mathbb{R}^{r\times n}_+$ such that $M\approx WH$. Note that $r$ is usually much smaller than $\min\{m,n\}$,  NMF is well-known as a powerful technique for dimension reduction,  and is able to give easily interpretable factors due to the nonnegativity constraints. It has been applied successfully in many areas, like image processing, text data mining, hyperspectral unmixing, see for example the recent survey and books \cite{fu2019nonnegative,gillis2020nonnegative,cichocki2009nonnegative} and the references therein.

In general, NMF is NP-hard and its solution is not unique, see \cite{vavasis2009complexity,fu2019nonnegative} and the reference therein. To resolve these two disadvantages, some assumptions like separability are introduced as a way to solve NMF problem efficiently and to guarantee the uniqueness of solution. NMF with separability assumption is referred to as separable NMF problem, aims to find nonnegative matrices $W\in \mathbb{R}^{m\times r}_+$ and $H\in \mathbb{R}^{r\times n}_+$ such that
$$
M=WH,\quad W=M(:,\mathcal{K}).
$$
The constraint $W=M(:,\mathcal{K})$ implies that each column of $W$ is equal to a column of $M$. 
If a matrix $M$ is $r$-separable, then there exist some permutation matrix $\Pi\in \{0,1\}^{n\times n}$ and a nonnegative matrix $H'\in \mathbb{R}^{r\times (n-r)}_+$ such that
\[
M\Pi=M\Pi\left(
           \begin{array}{cc}
             I_r  & H'  \\
              0_{n-r,r} & 0_{n-r,n-r}  \\
           \end{array}
         \right),
\]
where $I_r$ is the $r$-by-$r$ identity matrix and $0_{r,p}$ is the matrix of all zeros of dimension $r$ by $p$.
Equivalently, 
\begin{equation} \label{eq:convsep}
M \; = \; M \; \underbrace{ \Pi \left(
           \begin{array}{cc}
             I_r  & H'  \\
              0_{n-r,r} & 0_{n-r,n-r}  \\
           \end{array}
         \right) \Pi^{T} }_{X \in \mathbb{R}^{n \times n}}.
\end{equation}
This equivalent definition of separability was   proposed and discussed in~\cite{esser2012convex, recht2012factoring, elhamifar2012see, gillis2014robust} and will be very useful in this paper. 


 
%

Separable NMF is an important method which corresponding to self-dictionary learning in data science \cite{gillis2018fast}. The separability makes sense in many practical applications. For instance, in document classification, given a word-document data matrix, each entry $M(i,j)$ of $M$ represents the importance of word $i$ in document $ j $. Separability of $M$ indicates that, for each topic, there exist at least one document only discuss that topic, which is referred to as ''pure'' document. These ''pure'' documents can be regarded as key features that form feature matrix $W=M(:,\mathcal{K})$ to represent its original data matrix $M$. Considering feature matrix $W$, i.e., ''word $\times$ key documents'' matrix, it is reasonable to assume that , for each pure document, there are at least one word used only in that document. For example, in a pure document that only discusses biology, the words like ''transaminase'', ''amino acid'', can only show up in that biology document, but not in documents related to politics, philosophy or art.

Based on the above consideration, we generalize the separability assumption as follows.

\begin{defn}\label{def1}
A matrix $M\in \mathbb{R}^{m\times n}_+$ is co-$(r_1,r_2)$ separable if there exists an index set $\mathcal{K}_1$ of cardinality $r_1$ and an index set $\mathcal{K}_2$ of cardinality $r_2$, and nonnegative matrices $P_1\in \mathbb{R}^{m\times r_1}_+$ and $P_2\in \mathbb{R}^{r_2\times n}_+$ such that
\begin{equation}\label{cosep}
M=P_1 M(\mathcal{K}_1,\mathcal{K}_2) P_2
\end{equation}
where $P_1(\mathcal{K}_1,:)=I_{r_1}$ and $P_2(:,\mathcal{K}_2)=I_{r_2}$. $M(\mathcal{K}_1,\mathcal{K}_2)$ is referred to as the core of matrix $M$.  
\end{defn}
For simplicity, we call a matrix CoS-matrix if it has decomposition (\ref{cosep}). 
The co-$(r_1,r_2)$-separability is a natural extension of $r$-separability. A matrix $M$ is $r$-separable matrix, is also a co-$(m,r)$-separable. Every $m\times n$ nonnegative matrix $M$ is co-$(m,n)$-separable. 
Note-worthily, compared to $r$-separability, co-$(r_1,r_2)$-separability provides a more compact basic matrix (i.e., $M(\mathcal{K}_1,\mathcal{K}_2)$)
to represent data matrix. 


\subsection{Related Problems}

As a method that selects columns and rows  to represent the input nonnegative matrix , CoS-NMF model is related to generalized separable NMF (GS-NMF) model \cite{pan2019generalized}. Precisely, GS-NMF aims to find row set $\mathcal{K}_1$ and column set  $\mathcal{K}_2$ to represent $M$ in the form of $M=M(:,\mathcal{K}_2)P_2+P_1M(\mathcal{K}_1,:)$,
 where 
$P_2(:,\mathcal{K}_2)=I_{r_2}$ and $P_1(\mathcal{K}_1,:)=I_{r_1}$.
The motivation of GS-NMF is different from CoS-NMF model, take document classification as an example, GS-NMF assumes that there exists either a ''pure'' document or an anchor word, while CoS-NMF assume that there are at least an anchor word exist in a ''pure'' document. The different motivations lead to different representative form. We can see that GS-NMF is more relaxed, while CoS-NMF has a more compact form.  

CoS-NMF also has a very close connection with CUR decomposition,
that is, given a matrix $M$, identify a row subset $\mathcal{K}_1$ and column subset $\mathcal{K}_2$  from $M$ such that $\|M-M(:,\mathcal{K}_2)UM( \mathcal{K}_1,:)\|$ is minimized. For CUR model, the factor matrix $U$ is computed to minimize the approximation error \cite{mahoney2009cur}, i.e., $U=M(:,\mathcal{K}_2)^\dagger M M(\mathcal{K}_1,:)^\dagger$. When $U$ is required to be $U=M(\mathcal{K}_1,\mathcal{K}_2)^\dagger$, the variant model is then called pseudo-skeleton approximation where $A^\dagger$ denotes a Moore-Penrose generalized inverse of matrix $A$. Note that these models do not consider nonnegativity, the analysis is different from CoS-NMF. For example, CUR can pick any subset of $r$ linearly independent rows and columns to obtain exact decompositions of any rank-r matrix, but it is not true for CoS-NMF. For more information on CUR decomposition and pseudo-skeleton approximation, we refer the interest reader to  \cite{goreinov1997theory, mikhalev2018rectangular, wang2013improving,cai2021robust} and the references therein. In Section 3, we will discuss the connection and difference between CoS-NMF and CUR in details.

Our model is also related to tri-symNMF model proposed in \cite{arora2013practical} for learning topic models, i.e., given a word-document matrix $M$, it aims to find the word-topic matrix $W$ and topic-topic $S$ such that $A=MM^T\approx WSW^T$, where $A$ is the word co-occurrence matrix. Gillis in \cite{gillis2020nonnegative} showed that tri-symNMF model, can be represented in the form of $A=Wdiag(z)^{-1}A(\mathcal{K},\mathcal{K})diag(z)^{-1}W^T$, where $W(\mathcal{K},:)=diag(z)$ for some $z\in \mathbb{R}^r_+$.  Any separable NMF algorithms like SPA, can be hired to solve this model. One can solve a minimum-volume tri-symNMF instead since the separability assumption in tri-symNMF can  be relaxed to sufficiently scattered condition (SSC), see\cite{fu2019nonnegative,fu2018anchor} for more details. We note that if $diag(z)=I_r$,  tri-symNMF model is a special case of CoS-NMF, provided that $P_1=P^T_2$ in (\ref{cosep}).

In \cite{ding2006orthogonal,wang2011nonnegative}, Ding and et al. proposed a nonnegative matrix tri-factorization for co-clustering, i.e., given a matrix $M\in \mathbb{R}^{m\times n}_+$, it aims to find $G_1\in \mathbb{R}^{m\times r_1}_+$, $S\in \mathbb{R}^{r_1\times r_2}_+$ and $G_2\in \mathbb{R}^{r_2\times n}_+$ such that $M\approx G_1SG_2$.
It provides a good framework to simultaneously cluster the rows and columns of $M$. Here $G_1$ gives row clusters and $G_2$ gives column clusters.
When orthogonality constrain is added to $G_1$ and $G_2$, i.e., $G_1^TG_1=I$, $G_2G^T_2=I$, the model is called bi-orthogonal tri-factorization (BiOR-NM3F) and related to hard co-clustering. In Section 3, we will show the connection between  BiOR-NM3F and CoS-NMF.

\subsection{The Outline}

In this paper, we consider CoS-NMF problem which generates separability condition to co-separability on NMF problem.

In Section 2, some equivalent characterizations of CoS-matrix are first provided that lead to an ideal model to tackle CoS-NMF problem. We present some properties and discuss the uniqueness of CoS-NMF problem. We show that a minimal co-$(r_1,r_2)$-separable is unique up to scaling and permutations, while its selection of row set $\mathcal{K}_1$ and column set $\mathcal{K}_2$ is not unique.

In Section 3, we discuss the relationship between CoS-NMF problem and three other related problems. First, we give the intersection form of CoS-matrix and GS-matrix. Then, we present the relation with bi-orthogonal tri factorization (BiOR-NM3F) and prove that any matrix $M$ admits BiOR-NM3F form is minimal co-$(r_1,r_2)$-separable matrix. At last, we show the connection with CUR decomposition, that is, any CoS-matrix admits an exact CUR decomposition, while CUR-matrix is a CoS-matrix only under some conditions.

In Section 4, based on the properties of CoS-NMF, we propose a convex optimization model. An alternating fast gradient method generalized from the method presented in \cite{gillis2018fast}  is  proposed to tackle CoS-NMF problem.  

In Section 5, numerical experiments are conducted on synthetic , document and facial data sets. We show that CoS-NMF algorithm performs well in co-clustering applications, as well as preserves a good approximation to its original data matrix.

\section{Properties of Co-Separable Matrices}

In the following, we first present three equivalent characterizations of CoS matrices.

\begin{property}
[Equivalent Characterization 1]\label{prop:eqiv1}
 A matrix $M\in \mathbb{R}^{m\times n}_+$ is co-$(r_1,r_2)$-separable if and only if it can be written as 
\begin{equation} \label{M}
 M=\Pi_r\left(
              \begin{array}{cc}
                 S  & SH \\
                 WS & WSH \\
              \end{array}
            \right)\Pi_c,
 \end{equation}
  for some permutations matrices $\Pi_c \in \{0,1\}^{n\times n}$ and $\Pi_r \in \{0,1\}^{m\times m}$, and for some nonnegative matrices
  $S\in \mathbb{R}^{r_1 \times r_2}_+$,$W \in \mathbb{R}^{(m-r_1)  \times r_1}_+$
,
and
   $H \in \mathbb{R}^{r_2\times (n-r_2)}_+$.
\end{property}

\begin{proof}
The permutation $\Pi_r$ is chosen such that it moves the rows of $M$ corresponding to $\mathcal{K}_1$ in the first $r_1$ positions, and the permutation $\Pi_c$ is chosen such that it moves the columns of $M$ corresponding to $\mathcal{K}_2$ in the first $r_2$ positions. After the permutations, $M(\mathcal{K}_1,\mathcal{K}_2)$ is the $r_1$ by $r_2$ block in top left of $\Pi^T_rM\Pi^T_c$. Since  $M=P_1M(\mathcal{K}_1, \mathcal{K}_2)P_2$ for nonnegative matrices $P_1$ and $P_2$, we let $M(\mathcal{K}_1,\mathcal{K}_2)=S$, $P_1=\Pi_r\left(
              \begin{array}{c}
                 I_{r_1}  \\
                 W\\
              \end{array}
            \right)$, $P_2=[I_{r_2},H]\Pi_c$, the results follow.
\end{proof}

We know that a matrix $M$ is $r$-separable if and only if it can be written in the form of \eqref{eq:convsep}. In the following, we present a similar characterization for CoS matrices.

\begin{property}
[Equivalent Characterization 2] \label{prop:eqiv2}
A matrix $M\in \mathbb{R}^{m\times n}_+$
is co-$(r_1,r_2)$-separable if and only if it can be written as
\begin{equation}\label{eqiv2}
M=XMY,
\end{equation}
where
\begin{align}
X =\Pi_r\left(
             \begin{array}{cc}
               I_{r_1} &0_{r_1,m-r_1} \\
               W &0_{m-r_1,m-r_1} \\
             \end{array}
           \right)\Pi^T_r,  ~~       
           Y = \Pi^T_c\left(
             \begin{array}{cc}
             I_{r_2} &  H \\
               0_{n-r_2,r_2} & 0_{n-r_2,n-r_2}  \\
             \end{array}
           \right)\Pi_c, \label{eq:XY}
\end{align}
for some permutations matrices $\Pi_c \in \{0,1\}^{n\times n}$ and $\Pi_r \in \{0,1\}^{m\times m}$, and for some $W \in \mathbb{R}^{(m-r_1)\times r_1}_+$ and $H \in \mathbb{R}^{ r_2\times(n-r_2)}_+$.
\end{property}

\begin{proof}
From Property \ref{prop:eqiv1}, the matrix $M$ is co-$(r_1,r_2)$-separable if and only if there exist some permutation matrices $\Pi_c \in \{0,1\}^{n\times n}$ and $\Pi_r \in \{0,1\}^{m\times m}$ such that 
$$
 M=\Pi_r\left(
              \begin{array}{cc}
                 S  & SH \\
                 WS & WSH \\
              \end{array}
            \right)\Pi_c
$$
for some $W \in \mathbb{R}^{(m-r_1)\times r_1}_+$ and $H \in \mathbb{R}^{ r_2\times(n-r_2)}_+$. Let $\hat{M}=\Pi^T_r M \Pi^T_c= \left(
              \begin{array}{cc}
                 S  & SH \\
                 WS & WSH \\
              \end{array}
            \right)$, we have that 
$$
\hat{M}=\left(
             \begin{array}{cc}
               I_{r_1} &0_{r_1,m-r_1} \\
               W &0_{m-r_1,m-r_1} \\
             \end{array}
           \right)\hat{M}\left(
             \begin{array}{cc}
             I_{r_2} &  H \\
               0_{n-r_2,r_2} & 0_{n-r_2,n-r_2}  \\
             \end{array}
           \right).
$$
Since  $M=\Pi_r \hat{M} \Pi_c$, the result follows.
\end{proof}

Intuitively, co-separable should be intrinsically related to separable decomposition. In the following, we show a characterization that is represented by separability of matrix $M$ and its transpose $M^T$.

\begin{property}
[Equivalent Characterization 3] \label{prop:eqiv3}
A matrix $M\in \mathbb{R}^{m\times n}_+$
is co-$(r_1,r_2)$-separable if and only if  $M$ is $r_2$-separable and $M^T$ is 
$r_1$-separable, i.e., it can be written as
\begin{equation}\label{XMY}
M=XM,\quad M=MY
\end{equation}
where
\begin{align*}
X =\Pi_r\left(
             \begin{array}{cc}
               I_{r_1} &0_{r_1,m-r_1} \\
               W &0_{m-r_1,m-r_1} \\
             \end{array}
           \right)\Pi^T_r,  ~~
           Y  = \Pi^T_c\left(
             \begin{array}{cc}
             I_{r_2} &  H \\
               0_{n-r_2,r_2} & 0_{n-r_2,n-r_2}  \\
             \end{array}
           \right)\Pi_c, 
\end{align*}
for some permutations matrices $\Pi_c \in \{0,1\}^{n\times n}$ and $\Pi_r \in \{0,1\}^{m\times m}$, and for some $W \in \mathbb{R}^{(m-r_1)\times r_1}_+$ and $H \in \mathbb{R}^{ r_2\times(n-r_2)}_+$.
\end{property}

\begin{proof}
From Property \ref{prop:eqiv1}, after permutations $\Pi_c$ and $\Pi_r$, the last $n-r_2$ columns are convex combinations of the first $r_2$ columns of $\Pi^T_rM\Pi^T_c$, which implies that 
$M$ is $r_2$-separable; also the last $m-r_1$ rows are convex combinations of the first $r_1$ rows of $\Pi^T_rM\Pi^T_c$, which implies that 
$M^T$ is $r_1$-separable, i.e., (\ref{XMY}) established. 

Letting $\mathcal{M}=\{1,\cdots, m\}$, and $\mathcal{N}=\{1,\cdots, n\}$, if $M$ is $r_2$ separable, there exists a column set $\mathcal{K}_2$  such that $M=M(:,\mathcal{K}_2)Y_0$, where $Y_0\in \mathbb{R}^{r_2\times n}_+$ , $Y_0(:,\mathcal{K}_2)=I_{r_2}$ and $Y_0(:,\mathcal{N}-\mathcal{K}_2)=H$. Similarly, $M^T$ is $r_1$ separable, there exists a row set $\mathcal{K}_1$  such that $M=X_0M(\mathcal{K}_1,:)$, where $X_0\in \mathbb{R}^{m\times r_1}_+$, $X_0(\mathcal{K}_1, :)=I_{r_1}$ and $X_0(\mathcal{M}-\mathcal{K}_1,:)=W$. Letting
$S=M(\mathcal{K}_1,\mathcal{K}_2)$, we have,
$$
M(\mathcal{M}-\mathcal{K}_1,\mathcal{K}_2)=X_0 (\mathcal{M}-\mathcal{K}_1,\mathcal{K}_1)M(\mathcal{K}_1,\mathcal{K}_2)=WS,
$$
$$
M(\mathcal{K}_1,\mathcal{N}-\mathcal{K}_2)= M(\mathcal{K}_1,\mathcal{K}_2)Y_0(\mathcal{K}_2,\mathcal{N}-\mathcal{K}_2)=SH,
$$
$$
M(\mathcal{M}-\mathcal{K}_1,\mathcal{N}-\mathcal{K}_2)=X_0 (\mathcal{M}-\mathcal{K}_1,\mathcal{K}_1)M(\mathcal{K}_1,\mathcal{N}-\mathcal{K}_2)=WSH.
$$
Hence, after some permutations, $M$ has the form of (\ref{M}), i.e., $M$ is co-$(r_1,r_2)$-separable matrix.
\end{proof}

\begin{remark}\label{remark:1} For a co-$(r_1,r_2)$-separable matrix $M$, $M(:,\mathcal{K}_2)^T$ is $r_1$-separable, and $M(\mathcal{K}_1,:)$ is $r_2$-separable.
\end{remark}

In the following property, we will show that under some conditions,  a 
co-$(r_1,r_2)$-separable 
matrix can be further decomposed into a more compressive form.

\begin{property}\label{prop:compressive}
A co-$(r_1,r_2)$-separable matrix  
$
 M=\Pi_r\left(
              \begin{array}{cc}
                 S  & SH \\
                 WS & WSH \\
              \end{array}
            \right)\Pi_c \in \mathbb{R}^{m\times n}_+
$ 
can be reduced to 
co-$(\hat{r}_1,\hat{r}_2)$-separable matrix if 
the core $S\in \mathbb{R}^{r_1\times r_2}_+$ is co-$(\hat{r}_1,\hat{r}_2)$-separable, where $\max\{\hat{r}_1,\hat{r}_2\}\leq\min\{r_1,r_2\}$. 
\end{property}

\begin{proof}
$S$ is co-$(\hat{r}_1,\hat{r}_2)$-separable, without loss of generality, let 
$S=
\left(
              \begin{array}{c}
                 I_{\hat{r}_1}  \\ W_0\\
                 \end{array}
            \right)S_0
\left(
              \begin{array}{cc}
                 I_{\hat{r}_2}  & H_0\\
                 \end{array}
            \right)$, with $S_0\in \mathbb{R}^{\hat{r}_1\times \hat{r}_2}$, $W_0\in \mathbb{R}^{(r_1-\hat{r}_1)\times \hat{r}_1}$, $H_0\in \mathbb{R}^{\hat{r}_2\times (r_2-\hat{r}_2) }$, letting $W=[W_1,W_2]$,  $H=[H_1,H_2]^T$, where $W_1\in \mathbb{R}^{(m-r_1)\times \hat{r}_1}$, $W_2\in \mathbb{R}^{(m-r_1)\times (r_1-\hat{r}_1)}$, $H_1\in \mathbb{R}^{\hat{r}_2\times (n-r_2)}$, $H_2\in \mathbb{R}^{(r_2-\hat{r}_2)\times (n-r_2)}$, then,
\begin{eqnarray*}
 M&=&\Pi_r \left(
              \begin{array}{c}
                 I_{r_1}  \\
                 W
                 \end{array}
            \right)S
            \left(
              \begin{array}{cc}
                 I_{r_2} & H \\  
                 \end{array}
            \right)\Pi_c\\
  &=&\Pi_r\left(
              \begin{array}{cc}
                 I_{\hat{r}_1} & 0 \\ 0&I_{r_1-\hat{r}_1}\\
                 W_1& W_2
                 \end{array}
            \right)\left(
              \begin{array}{c}
                 I_{\hat{r}_1}  \\ W_0\\
                 \end{array}
            \right)S_0
\left(
              \begin{array}{cc}
                 I_{\hat{r}_2}  & H_0\\
                 \end{array}
            \right)\left(
              \begin{array}{ccc}
                 I_{\hat{r}_2} & 0 & H_1 \\ 0&I_{r_2-\hat{r}_2}&H_2
                 \end{array}
            \right)\Pi_c \\
            &=& \Pi_r\left(
              \begin{array}{c}
                 I_{\hat{r}_1}  \\ W_0\\W_1+W_2W_0\\
                 \end{array}
            \right)S_0            
\left(
              \begin{array}{ccc}
                 I_{\hat{r}_2}  & H_0& H_1+H_0H_2\\
                 \end{array}
            \right)\Pi_c.
\end{eqnarray*}
Hence, $M$ is reduced to co-$(\hat{r}_1,\hat{r}_2)$-separable. The results follow. 
\end{proof}

Although a co-$(r_1,r_2)$-separable matrix can be further compressed from Property \ref{prop:compressive}, we need to remark the following fact to the authors.

\begin{remark}\label{remark2}
A co-$(r_1,r_2)$-separable matrix $M$ is not always a co-$(r,r)$-separable, where $r=\min\{r_1,r_2\}$.
\end{remark}

\begin{example}
Given $M=[m_1,m_2,\cdots,m_r,m_{r+1}]\in \mathbb{R}^{r\times (r+1)}_+$,  and $[m_1,m_2,\cdots,m_r]\in \mathbb{R}^{r\times r}_+$ has full column rank $r$ and $m_{r+1}=\alpha_1 m_1+\cdots+\alpha_{r-1} m_{r-1}-\alpha_r m_r$, where $\alpha_i\geq 0$,  $i=1,\cdots, r$. We could not find a column set $W\in \mathbb{R}^{r\times r}\subset M$, such that  $M=W[I_r,h]$ and $h\geq 0$, 
i.e., $M$ is not co-$(r,r)$-separable, but co-$(r,r+1)$-separable. 
\end{example}

From Property \ref{prop:compressive}, given a co-$(r_1,r_2)$-separable matrix $M$, it becomes important to find the minimal value for $r_1$ and $r_2$  since this compresses the data matrix the most. In following, we define minimal co-$(r_1, r_2)$-separable matrices. 

 \begin{defn}\label{defn:minisep}
 A matrix $M$ is a minimal co-$(r_1,r_2)$-separable if $M$ is co-$(r_1,r_2)$-separable and $M$ is not co-$(r'_1,r'_2)$-separable for any $r_1'< r_1$ and $r_2'<r_2$.
 \end{defn}

In general, from Remark \ref{remark2}, $r_1$ is not necessary equal to $r_2$ for a minimal co-$(r_1,r_2)$-separable matrix $M$. However, there are some exceptions. The simplest cases are for rank-one and rank-two matrices.

\begin{property}
Any nonnegative rank one matrix $M$  is minimal co-$(1,1)$-separable matrix.
\end{property}

\begin{proof}
It follows directly from the fact that all the  rows (resp$.$ columns) are multiple of one another.
\end{proof}

\begin{property}
Any nonnegative rank two matrix $M$  is minimal co-$(2,2)$-separable matrix.
\end{property}

\begin{proof}
We know that any two dimensional cone can be always spanned by its two extreme rays, and $rank(M)=rank(M^T)$, it means both $M$ and $M^T$ are 2-separable. From Property \ref{prop:eqiv3}, $M$ is co-$(2,2)$-separable.
\end{proof}

For a general case, from Property \ref{prop:eqiv3}, finding minimal CoS factorization is 
equivalent to finding X and Y that satisfy \eqref{XMY} and such that
the number of non-zero rows of X and non-zero columns of
Y is minimized.

 \begin{property}[Idealized Model]\label{Pro:idealmodel} Let $M$ be minimal 
 co-$(r_1,r_2)$-separable, and let $(X^*,Y^*)$ be an optimal solution of
\begin{equation}\label{eq:idealmodel}
\begin{split}
 \min_{X\in \mathbb{R}^{m\times m}_+,Y\in \mathbb{R}^{n\times n}_+}    \|X\|_{col,0}+\|Y\|_{row,0},\quad
  \mbox{ s.t. } \; M=XM, \quad M=MY.
 \end{split}
 \end{equation}
$ \|X\|_{col,0}$ equal to the number  of nonzero columns of $X$ and $\|Y\|_{row,0}$ equals to the number of nonzero rows of $Y$.
Let also $\mathcal{K}_1$ correspond to the indices of the non-zero columns of $X^*$ and
$\mathcal{K}_2$ to the indices of the non-zero rows of $Y^*$, then $|\mathcal{K}_1|+|\mathcal{K}_2|=r_1+r_2$.
\end{property}

\begin{proof}
On one hand, $M$ is minimal co-$(r_1,r_2)$-separable matrix, from Property \ref{prop:eqiv3}, there exist $X$ and $Y$ such that $M=XM$ and $M=MY$, where the number of nonzero column of $X$ and  nonzero row of $Y$ is equal to $r_1+r_2$. By the optimality of $(X^*, Y^*)$, we have $|\mathcal{K}_1|+|\mathcal{K}_2|\leq r_1+r_2.$

On the other hand, without loss of generality, let the optimal solution $(X^*,Y^*)$ and matrix $M$ be
$$
X^*=\left(
              \begin{array}{cc}
                 X_1  & 0 \\
                 X_2 & 0 \\
              \end{array}
            \right),~~
             Y^*=\left(
              \begin{array}{cc}
                 Y_1  & Y_2 \\
                 0 & 0 \\
              \end{array}
            \right),~~
      M=\left(
              \begin{array}{cc}
                 M_{11}  & M_{12} \\
                 M_{21} & M_{22} \\
              \end{array}
            \right).
$$
At least one principal submatrix of $X$ of order $|\mathcal{K}_1|$ is nonsingular, without loss of generality, let $X_1\in \mathbb{R}^{|\mathcal{K}_1|\times |\mathcal{K}_1| }$ be full rank submatrix of $X$. Similarly, let $Y_1\in\mathbb{R}^{|\mathcal{K}_2|\times |\mathcal{K}_2| }$ be full rank submatrix of $Y$. From $M=XM$ and $M=MY$, we have
$$
\left(
              \begin{array}{cc}
                 M_{11}  & M_{12} \\
                 M_{21} & M_{22} \\
              \end{array}
            \right)=\left(
              \begin{array}{cc}
                 X_1M_{11} & X_1 M_{12}  \\
                 X_2M_{11} &  X_2 M_{12}  \\
              \end{array}
            \right),~~
            \left(
              \begin{array}{cc}
                 M_{11}  & M_{12} \\
                 M_{21} & M_{22} \\
              \end{array}
            \right)=\left(
              \begin{array}{cc}
                 M_{11}Y_1 & M_{11}Y_2 \\
                M_{21}Y_1 &  M_{21}Y_2 \\
              \end{array}
            \right),  
$$
hence, 
$M_{12}=M_{11}Y_2$, $M_{21}=X_2M_{11}$, $M_{22}=X_2M_{12}=X_2(M_{11}Y_2)$.
We have,  
\begin{eqnarray*}
M=\left(    \begin{array}{cc}
                 M_{11}  &   M_{11}Y_2 \\
                X_2M_{11} & X_2M_{11}Y_2 \\
              \end{array}
            \right) 
            =\left(
              \begin{array}{c}
                 I_{|\mathcal{K}_1|}  \\
                X_2    \\
              \end{array}
            \right) 
                 M_{11} \left(
              \begin{array}{cc}
                I_{|\mathcal{K}_2|}  &  Y_2 \\
              \end{array}
            \right),
\end{eqnarray*} 
from Definition \ref{def1},  we know that $M$ is 
co-$(|\mathcal{K}_1|,|\mathcal{K}_2|)$-separable matrix. Since $M$ 
is a minimal-$(r_1,r_2)$-separable, from  Definition \ref{defn:minisep}, 
$|\mathcal{K}_1|+|\mathcal{K}_2|\geq r_1+r_2$.
 Therefore, the result follows. 
\end{proof}

The following property shows that co-$(r_1,r_2)$-separability is invariant to scaling.

\begin{property}\label{scaling}[Scaling]
The matrix $M$ is co-$(r_1,r_2)$-separable if and only if $D_1 MD_2$ is 
$(r_1,r_2)$-separable for any diagonal matrices $D_1$ and $D_2$ whose diagonal elements are positive.
\end{property}

\begin{proof}
Let $M$ be co-$(r_1,r_2)$ separable with $M=P_1M(\mathcal{K}_1,\mathcal{K}_2)P_2$ with $|\mathcal{K}_1|=r_1$ and $|\mathcal{K}_2|=r_2$. Multiplying on both sides by $D_1$ and $D_2$, we have
\begin{eqnarray*}
D_1MD_2&=&D_1P_1M(\mathcal{K}_1,\mathcal{K}_2)P_2D_2 \\
&=&D_1P_1D^{-1}_1(\mathcal{K}_1,\mathcal{K}_1)\big(D_1(\mathcal{K}_1,\mathcal{K}_1)M(\mathcal{K}_1,\mathcal{K}_2)D_2(\mathcal{K}_2,\mathcal{K}_2)\big)D^{-1}_2(\mathcal{K}_2,\mathcal{K}_2) P_2 D_2.
\end{eqnarray*}
Denoting $\tilde{M}=D_1 M D_2$, $\tilde{P}_1=D_1 P_1 D^{-1}_1(\mathcal{K}_1, \mathcal{K}_1)$, $\tilde{P}_2=D^{-1}_2(\mathcal{K}_2, \mathcal{K}_2) P_2 D_2$, note that
\begin{eqnarray*}
\tilde{M}(\mathcal{K}_1, \mathcal{K}_2)&=&D_1(\mathcal{K}_1,:)MD_2(:,\mathcal{K}_2)\\
&=& D_1(\mathcal{K}_1,:)P_1M(\mathcal{K}_1,\mathcal{K}_2)P_2D_2(:,\mathcal{K}_2)\\
&=& D_1(\mathcal{K}_1,\mathcal{K}_1) M(\mathcal{K}_1,\mathcal{K}_2) D_2(\mathcal{K}_2,\mathcal{K}_2),
\end{eqnarray*}
therefore, we have
$$
\tilde{M}=\tilde{P}_1\tilde{M}(\mathcal{K}_1,\mathcal{K}_2)\tilde{P}_2.
$$
Moreover, $\tilde{P}_1(\mathcal{K}_1,:)=I_{r_1}$ and $\tilde{P}_2(:,\mathcal{K}_2)=I_{r_2}$; hence 
$\tilde{M}$ is co-$(r_1,r_2)$-separable. The proof of the other direction is the same because $\tilde{M}=D_1MD_2$ is the diagonal scaling of $M$ using 
the inverses of $D_1$ and $D_2$.
\end{proof}

\subsection{Uniqueness of CoS-NMF}


%

Similar to separable NMF, CoS-NMF admits a unique solution up to permutation and scaling.

\begin{property}[Uniqueness]\label{Pro:unique}
Let $M\in \mathbb{R}^{m\times n}_+$ be a minimal co-$(r_1,r_2)$-separable matrix
in the form of (\ref{cosep}) and $rank(M)=r_1=r_2$, then $(P_1,S,P_2)$ is unique  up to permutation and scaling, i.e., there exists permutation matrices $\Pi_1\in \{0,1\}^{r_1\times r_1}$, $\Pi_2\in \{0,1\}^{r_2\times r_2}$,  and diagonal scaling matrix $D_1\in \mathbb{R}^{r_1\times r_1}$ and $D_2\in \mathbb{R}^{r_2\times r_2}$ with positive elements such that 
$$
\tilde{P}_1=P_1\Pi_1D^{-1}_1, \quad\tilde{S}=D_1\Pi^T_1S\Pi^T_2D_2,\quad \tilde{P}_2=D^{-1}_2\Pi_2P_2.
$$
\end{property}

\begin{proof}
$M$ is co-$(r_1,r_2)$-separable matrix, i.e., $M=P_1SP_2$, where $P_1(\mathcal{K}_1,:)=I_{r_1}$ and $P_2(:,\mathcal{K}_2)=I_{r_2}$ are separable matrices.
From Lemma 4.36 and Theorem 4.37 in \cite{gillis2020nonnegative}, we have
$$
cone(M)=cone(P_1S),\quad cone(M^T)=cone(P^T_2S^T).
$$
When $r_1=r_2=r$, then $cone(M)$ has extreme rays which are columns of $P_1S$.  If there is another solution $(\tilde{P}_1,\tilde{S},\tilde{P}_2)$,  again  $cone(M)$ has extreme rays which are columns of $\tilde{P}_1\tilde{S}$, then columns of  $P_1S$ are  coincide up to scaling and permutation. Hence, $P_2$ is unique given $P_1S$. 
Similarly, $cone(M^T)$ has extreme rays which are columns of $P^T_2S^T$, i.e., the rows of $SP_2$. Hence, $P_1$ is unique, up to scaling and permutation. Since $P_1$ and $P_2$ are unique up to scaling and permutation, $S$ is unique. the results follow.
\end{proof}

Though from Property \ref{Pro:unique}, a minimal co-$(r,r)$-separable matrix is unique up to scaling and permutation, its selection of row set $\mathcal{K}_1$ and column set $\mathcal{K}_2$ is not unique. Here is an example.

\begin{example}
$$
M=\left(\begin{array}{ccc}
                SD_2  & S & SH_2 \\
                D_1SD_2 & D_1S& D_1SH_2 \\
                W_1SD_2& W_1S& W_1SH_2
              \end{array}
            \right),
            $$
where $S\in \mathbb{R}^{r\times r}$, $D_1$ and $D_2$ are  diagonal matrix with positive elements. We have
\begin{eqnarray*}
 M =\left(\begin{array}{c}
                I   \\
                D_1\\
                W_1
              \end{array}
            \right)S\left(\begin{array}{ccc}
                D_2  &  I& H_2
              \end{array}
            \right):~~
             M(\mathcal{K}_1, \mathcal{K}_2)=S,~~(\mathcal{K}_1, \mathcal{K}_2)=(1:r, r+1:2r)
\end{eqnarray*}
and 
$$
M =\left(\begin{array}{c}
                 D^{-1}_1  \\
                I\\
                W_1 D^{-1}_1
              \end{array}
            \right)D_1SD_2\left(\begin{array}{ccc}
                I & D^{-1}_2  & D^{-1}_2 H_2
              \end{array}
            \right):~~
              M(\mathcal{K}_1, \mathcal{K}_2) =D_1SD_2,~~(\mathcal{K}_1, \mathcal{K}_2)=(r+1:2r, 1:r).
$$
\end{example}
However, it is still possible to guarantee the uniqueness of the  selection of $(\mathcal{K}_1,\mathcal{K}_2)$. In the following, we propose some conditions for selection uniqueness.

\begin{property}
Let $M$ be minimal co-$(r_1,r_2)$-separable matrix, and $r_1=r_2=r$, there are no proportional columns and rows, then it admits a unique co-separable decomposition of size $(r,r)$.
\end{property}

\begin{proof}
The property is from Property \ref{Pro:unique} directly.
\end{proof}

\section{The Relationships with Other Matrix Factorizations}

A matrix $M\in \mathbb{R}^{m\times n}$ is $(l_1,l_2)$-generalized-separable 
matrix (GS-matrix\cite{pan2019generalized}) if  
\begin{equation}\label{Def:GSnmf}
M=M(:,\mathcal{L}_2)Z_2+Z_1M(\mathcal{L}_1,:),
\end{equation}
where 
$Z_2(:,\mathcal{L}_2)=I_{l_2}$ and $Z_1(\mathcal{L}_1,:)=I_{l_1}$. We remark that $M(\mathcal{L}_1,\mathcal{L}_2)=0$ for a $(l_1,l_2)$-GS-matrix  
\footnote{It is $(l_2,l_1)$-GS-matrix in \cite{pan2019generalized}, in order to keep the consistency of the subscript in this paper, we swapped row and column subscripts.}.
In the following property, we present the form of the intersection of 
co-$(r_1,r_2)$-separable matrix and $(l_1,l_2)$-GS-matrix.

\begin{property}[Relationship between GS-matrix]

If matrix $M$ has a unique minimal $(l_1,l_2)$-generalized-separable decomposition, and also admits  co-$(r_1,r_2)$-separable decomposition, then \\
$\min\{r_1,r_2\}\geq (l_1+l_2)$, and $M$ can be written as
\begin{equation}\label{prop6}
\small
M=\Pi_r\left(
             \begin{array}{ccc}
               Q_0W_0 &Q_0W_0H_1+Q_0W_1H_0+Q_1H_0&Q_0W_0U_1+Q_0W_0H_1U_0+
               Q_0W_1H_0U_0+Q_1H_0U_0\\
               W_0 &W_0H_1+W_1H_0&W_0U_1+W_0H_1U_0+W_1H_0U_0 \\
 0_{l_1,l_2}&H_0&H_0U_0              
             \end{array}
           \right)\Pi_c
\end{equation}
where $W_0\in \mathbb{R}^{(r_1-l_1)\times l_2}_+$, $W_1\in \mathbb{R}^{(r_1-l_1)\times l_1}_+$, $H_0\in \mathbb{R}^{l_1\times (r_2-l_2)}_+$, $H_1\in \mathbb{R}^{l_2\times (r_2-l_2)}_+$, $U_0\in \mathbb{R}^{(r_2-l_2)\times (n-r_2)}_+$, $U_1\in \mathbb{R}^{l_2\times (n-r_2)}_+$, $Q_0\in \mathbb{R}^{(m-r_1)\times (r_1-l_1)}_+$, $Q_1\in \mathbb{R}^{(m-r_1)\times l_1}_+$.
\end{property}

\begin{proof}
Matrix $M$ is co-$(r_1,r_2)$-separable, from Property \ref{prop:eqiv3}, $M$ is 
$r_1$-separable, i.e., $M$ is GS-$(r_1,0)$-separable. Since $M$ is 
minimal $(l_1,l_2)$-GS-separable, it implies that $(l_1+l_2)\leq r_1$. 
Similarly, $(l_1+l_2)\leq r_2$. Thus we have $\min\{r_1,r_2\}\geq (l_1+l_2)$.

$M$ is minimal $(l_1,l_2)$-GS-matrix, from (\ref{Def:GSnmf}), let $\mathcal{L}_2=\{l^{(2)}_1,\cdots,l^{(2)}_{l_2}\}$ and $\mathcal{L}_1=\{l^{(1)}_1,\cdots,l^{(1)}_{l_1}\}$ be the column and row set respectively, and $M(\mathcal{L}_1,\mathcal{L}_2)=0$. $M$ is co-$(r_1,r_2)$-separable matrix, from (\ref{cosep}), let $\mathcal{K}_2=\{k^{(2)}_1,\cdots,k^{(2)}_{r_2}\}$ and $\mathcal{K}_1=\{k^{(1)}_1,\cdots,k^{(1)}_{r_1}\}$ be the column and row set respectively in $ (\ref{cosep})$.
 
From Property 3, we have 
\begin{equation}\label{eqprop6}
M=M(:,\mathcal{K}_2)P_2, \quad M=P_1M(\mathcal{K}_1,:)
\end{equation}
where $P_1(\mathcal{K}_1,:)=I_{r_1}$ and $P_2(:,\mathcal{K}_2)=I_{r_2}$.
 
\textbf{Case 1}. If $\mathcal{L}_2\cap \mathcal{K}_2=\emptyset$,  then from (\ref{eqprop6}) and (\ref{Def:GSnmf}),
$$
M(:,\mathcal{L}_2)=M(:,k^{(2)}_1)P_2(1,\mathcal{L}_2)+\cdots+M(:,k^{(2)}_{r_2})P_2(r_2,\mathcal{L}_2)\neq 0
$$
and
\begin{eqnarray*}
M(\mathcal{L}_1,\mathcal{L}_2) =M(\mathcal{L}_1,k^{(2)}_1)P_2(1,\mathcal{L}_2)+\cdots+
 M(\mathcal{L}_1,k^{(2)}_{r_2})P_2(r_2,\mathcal{L}_2) =0.
\end{eqnarray*}
Therefore, we have
$M(\mathcal{L}_1,k^{(2)}_j)P_2(j,\mathcal{L}_2)=0$
for all $j\in \{1,\cdots, r_2\}$.

If $M(\mathcal{L}_1,k^{(2)}_j)\neq 0$ for all $j\in \{1,\cdots,r_2\}$, then
$P_2(j,\mathcal{L}_2)=0$, which implies $M(:,\mathcal{L}_2)=0$, contradicts to $M(:,\mathcal{L}_2)\neq 0$.

If $M(\mathcal{L}_1,k^{(2)}_j)= 0$ and $P_2(\mathcal{J},\mathcal{L}_2)\neq 0$ for some $j\in \mathcal{J}$, let $\bar{\mathcal{J}}\doteq\{1,\cdots,r_2\} - \mathcal{J}$ be the complement of $\mathcal{J}$ , we have $P_2(\bar{J},\mathcal{L}_2)= 0$, then
$M(:,\mathcal{L}_2)=M(:,\mathcal{K}_2(\mathcal{J}))P_2(\mathcal{J},\mathcal{L}_2)$. Hence, there exist $\tilde{Z}_2$ such that
$$
M=M(:,\mathcal{K}_2(\mathcal{J}))\tilde{Z}_2+Z_1M(\mathcal{L}_1,:),\quad \tilde{Z}_2(\mathcal{K}_2(\mathcal{J}),:)=I_{|\mathcal{J}|}.
$$
It contradicts to that $M$ has unique minimal $(l_1,l_2)$-generalized-separable decomposition. 
Hence, from the above analysis, we get $\mathcal{L}_2\cap \mathcal{K}_2\not =\emptyset$.
 
\textbf{Case 2}.
  If $\mathcal{L}_2\cap \mathcal{K}_2\neq\emptyset$, and $\mathcal{L}_2\not\subset \mathcal{K}_2$, let $\mathcal{L}^s_2\subset \mathcal{L}_2$ and $\mathcal{L}^s_2\not\subset \mathcal{K}_2$, from (\ref{eqprop6}) and (\ref{Def:GSnmf}),
$$
M(:,\mathcal{L}^s_2)=M(:,k^{(2)}_1)P_2(1,\mathcal{L}^s_2)+\cdots+M(:,k^{(2)}_{r_2})P_2(r_2,\mathcal{L}^s_2)\neq 0
$$
and
\begin{eqnarray*}
M(\mathcal{L}_1,\mathcal{L}^s_2) =M(\mathcal{L}_1,k^{(2)}_1)P_2(1,\mathcal{L}^s_2)+\cdots+
 M(\mathcal{L}_1,k^{(2)}_{r_2})P_2(r_2,\mathcal{L}^s_2)
=0,
\end{eqnarray*}
we have
$M(\mathcal{L}_1,k^{(2)}_j)P_2(j,\mathcal{L}^s_2)=0$
for all $j\in \{1,\cdots, r_2\}$. Similar to the proof in Case 1, we can prove that $\mathcal{L}_2\not\subset \mathcal{K}_2$ is not established. Therefore  $\mathcal{L}_2\subseteq \mathcal{K}_2$ is the only possibility.

Similarly, we can deduce that  $\mathcal{L}_1\subseteq \mathcal{K}_1$. It then leads the following discussion.
 
\textbf{Discussion}: When $\mathcal{L}_1\subseteq \mathcal{K}_1$ and  $\mathcal{L}_2\subseteq \mathcal{K}_2$, $M$ is $(l_1,l_2)$ GS-separable, we have
$$
M=\Pi_r\left(
              \begin{array}{cccc}
                M_{11}  & M_{12}&|&M_{13} \\
                M_{21} &M_{22}&|& M_{23} \\
                0_{l_1,l_2}&M_{32}&|& M_{33}
              \end{array}
            \right)\Pi_c \doteq  \Pi_r \hat{M} \Pi_c,
$$
where $\hat{M}=\Pi^T_rM\Pi^T_c$, $M_{11}\in \mathbb{R}^{(m-r_1)\times l_2}$, $M_{12}\in \mathbb{R}^{(m-r_1)\times (r_2-l_2)}$, $M_{13}\in \mathbb{R}^{(m-r_1)\times (n-r_2)}$, $M_{21}\in \mathbb{R}^{(r_1-l_1)\times (l_2)}$, $M_{22}\in \mathbb{R}^{(r_1-l_1)\times (r_2-l_2)}$, $M_{23}\in \mathbb{R}^{(r_1-l_1)\times (n-r_2)}$, $M_{32}\in \mathbb{R}^{l_1\times (r_2-l_2)}$, $M_{33}\in \mathbb{R}^{l_1\times (n-r_2)}$.  For simplicity, we will discuss $\hat{M}$ first. 

Since
$\hat{M}$ is co-$(r_1,r_2)$-separable, and $\hat{M}(m-r_1+1:m, 1:r_2)$ is the core.
From (\ref{eqprop6}), there exists $\hat{P}_2= \left(
              \begin{array}{ccc}
               I_{l_2}  & 0 &U_1 \\
               0& I_{r_2-l_2} & U_0
              \end{array}
            \right)$, $U_0\in \mathbb{R}^{(r_2-l_2)\times (n-r_2)}$, $U_1\in \mathbb{R}^{l_2\times (n-r_2)}$, such that $\hat{M}=\hat{M}(:,1:r_2)\hat{P}_2$, that is,
\begin{eqnarray*}
\hat{M}=\left(
              \begin{array}{cccc}
                M_{11}  & M_{12}&|&M_{11}U_1+M_{12}U_0 \\
                M_{21} &M_{22}&|& M_{21}U_1+M_{22}U_0 \\
                0_{l_1,l_2}&M_{32}&|& M_{32}U_0
              \end{array}
            \right)
\end{eqnarray*} 
Furthermore, there exists $\hat{P}_1= \left(
              \begin{array}{cc}
               Q_0 & Q_1\\
               I_{r_1-l_1}  & 0  \\
               0& I_{l_1} 
              \end{array}
            \right)$, $Q_0\in \mathbb{R}^{(m-r_1)\times (r_1-l_1)}$, $Q_1\in \mathbb{R}^{(m-r_1)\times l_1}$, such that $\hat{M}=\hat{P}_1\hat{M}(m-r_1+1:m,:)$, that is,
$M_{11}= Q_0M_{21}$, $M_{12}=Q_0M_{22}+Q_1M_{32}$, $M_{13}=Q_0M_{21}U_1+Q_0M_{22}U_0+Q_1M_{32}U_0$, $M_{23}=M_{21}U_1+M_{22}U_0 $.
          

Because $\hat{M}$ is $(l_1,l_2)$ GS-matrix, we deduce that there exist $W_1\in \mathbb{R}^{(r_1-l_1)\times l_1}$, $H_1\in \mathbb{R}^{l_2\times (r_2-l_2)}$ such that $M_{22}=M_{21}H_1+W_1M_{32}$, let $M_{21}=W_0$, $M_{32}=H_0$, we have (\ref{prop6}), the result follows. 
  \end{proof}

\begin{remark}
A $(r_1,0)$- (or $(0,r_2)$-) GS-matrix $M$ is also co-$(r_1,n)$- 
(or co-$(m ,r_2)$-) separable. 
\end{remark}

In the following, we will show an example that a $(l_1,l_2)$-GS-matrix $M\in \mathbb{R}^{m\times n}_+$ is not a co-$(r_1,r_2)$-separable matrix; and an example that co-$(r_1,r_2)$-separable matrix  $M\in \mathbb{R}^{m\times n}_+$ is not a 
$(l_1,l_2)$-GS-matrix, where $r_1<m, r_2<n$ and $(l_1+l_2)<\min\{m,n\}$.

\begin{example}
$M=\left(
              \begin{array}{cccc}
              1&2&8&7  \\
              2&2&8&7  \\
              2&1&7&5  \\
              0&0&2&1  \\
              \end{array}
            \right)= M(:,1:2)\left(
              \begin{array}{cccc}
             1&0 &2&1   \\
             0&1 &1&2
              \end{array}
            \right)+\left(
              \begin{array}{c}
             2 \\
             1\\
             1\\
             1\\
              \end{array}
            \right)M(4,:)$ is $(1,2)$-GS-matrix but not a co-$(3,3)$-separable matrix.
Let $M=\left(
              \begin{array}{c}
               I_{r_1}   \\
               W
              \end{array}
            \right)S\left(
              \begin{array}{cc}
               I_{r_2} & H
              \end{array}
            \right)$
where entries of $S\in \mathbb{R}^{r_1\times r_2}_+$, $W\in \mathbb{R}^{(m-r_1)\times r_1}_+$, $H\in \mathbb{R}^{r_2\times (n-r_2)}_+$ are positive. $M$ is 
co-$(r_1,r_2)$-separable matrix but not a $(l_1,l_2)$-GS matrix with $l_1+l_2<\min\{m,n\}$. 
\end{example}

The following property presents the relationship between CoS-matrix and a matrix which admits bi-orthogonal tri-factorization (BiOR-NM3F)\cite{ding2006orthogonal,wang2011nonnegative}, i.e.,
given a matrix $M\in \mathbb{R}^{m\times n}$, 
\begin{equation}\label{trinmf}
M=G_1SG_2,\quad G_1^TG_1=I_{r_1} , \quad   G_2G^T_2=I_{r_2},
\end{equation}
where $G_1\in \mathbb{R}^{m\times r_1}_+$, $S\in \mathbb{R}^{r_1\times r_2}_+$ and $G_2\in \mathbb{R}^{r_2\times n}_+$.  We call this type matrix $M$ a BiOR-NM3F matirx.  

\begin{property}[Relationship between BiOR-NM3F]
A BiOR-NM3F matrix $M$ is co-$(r_1,r_2)$-separable matrix.
\end{property}

\begin{proof}
From BiOR-NM3F setting,  $G_1$ is nonnegative orthogonal matrix, $G_1$ hence has only one positive entry in each row. Similarly, $G_2$ has only one positive entry in each column. Let $\alpha_i=\max_j G_1(j,i)$ and $\beta_i=\max_j G_2(i,j)$, $p_i$ be the number of nonzero entry of $i$-th column of $G_1$, $q_i$ be the number of nonzero entry of $i$-th row of $G_2$, and 
$$
\Lambda_1=\left(
              \begin{array}{ccc}
                D^{(1)}_1(\alpha_1) & \cdots &0 \\
               0 &\ddots & 0 \\
                0&\cdots& D^{(1)}_{r_1}({\alpha_{r_1})}
              \end{array}
            \right),$$
$$            
             \Lambda_2=\left(
              \begin{array}{ccc}
                D^{(2)}_1(\beta_1) & \cdots &0 \\
               0 &\ddots & 0 \\
                0&\cdots& D^{(2)}_{r_1}({\beta_{r_2})}
              \end{array}
            \right),
 $$
where $D^{(1)}_i(\alpha_i)\in \mathbb{R}^{p_i\times p_i}$  is diagonal matrix with $\alpha_i$ in the diagonal, $D^{(2)}_i(\beta_i)\in \mathbb{R}^{q_i\times q_i}$  is diagonal matrix with $\beta_i$ in the diagonal. Hence,
$$
M=\Lambda_1 \Lambda^{-1}_1 G_1 S G_2 \Lambda^{-1}_2 \Lambda_2 \Longrightarrow \Lambda^{-1}_1M\Lambda^{-1}_2= \Lambda^{-1}_1 G_1 S G_2 \Lambda^{-1}_2.
$$
Let $\tilde{M}=\Lambda^{-1}_1M\Lambda^{-1}_2$, $\tilde{P}_1=\Lambda^{-1}_1 G_1$, $\tilde{P}_2= G_2\Lambda^{-1}_2$,  we have $\tilde{P}_1$ contains an identity matrix $I_{r_1}$ and $\tilde{P}_2$ contains an identity matrix $I_{r_2}$. From co-separable definition \ref{cosep}, $\tilde{M}$ is a co-$(r_1,r_2)$-separable matrix. By Property \ref{scaling}, $M$ is co-$(r_1,r_2)$-separable. 
\end{proof}

As we mentioned in Section 1 that Co-separable NMF is also related to the CUR decomposition which identifies a row subset $\mathcal{K}_1$ and  column subset $\mathcal{K}_2$  from $M$ such that 
$\|M-M(:,\mathcal{K}_2)UM(\mathcal{K}_1,:)\|$ is minimized, where $|\mathcal{K}_1|=r_1$ and $|\mathcal{K}_2|=r_2$. For simplicity, we call a matrix the $(r_1,r_2)$-CUR matrix that admits an exact CUR decomposition. Different from Co-separable NMF,  nonnegativity constraints are not considered in CUR decomposition, which leads to a different model analysis. In the following property, we will show  a connection between both models.

\begin{property}[Relationship between CUR] 
(i)
Any minimal co-$(r_1,r_2)$-separable matrix $M$ admits an exact CUR decomposition. 
(ii) If a nonnegative matrix $M$ admits an exact CUR
decomposition, i.e.,  $M=M(:,\mathcal{K}_2)UM(\mathcal{K}_1,:)$, and  $U\geq 0$, then $M$ is minimal co-$(r_1,r_2)$-separable matrix with core $M (\mathcal{K}_1,\mathcal{K}_2)$.
\end{property}

\begin{proof}
$M$ is minimal co-$(r_1,r_2)$-separable, from property \ref{prop:eqiv1}, we have
\begin{eqnarray*} \label{prop_cur}
 M&=&\Pi_r\left(
              \begin{array}{cc}
                 S  & SH \\
                 WS & WSH \\
              \end{array}
            \right)\Pi_c \\
    &=& \Pi_r\left(
              \begin{array}{c}
                 S   \\
                 WS  \\
              \end{array}
            \right)\Pi_c  (\Pi^T_c S^{+} \Pi^T_r)  \Pi_r\left(
              \begin{array}{cc}
                 S  & SH \\
              \end{array}
            \right)\Pi_c \\
            &=& M(:,\mathcal{K}_2)  (\Pi^T_c S^{+} \Pi^T_r)  M(\mathcal{K}_1,:).
             \end{eqnarray*}
where $S^+$ is referred to as moore-penrose inverse of $S$.  $M$ admits an exact CUR decomposition.

If $M$ admits an exact CUR decomposition, i.e., $M=M(:,\mathcal{K}_2)UM(\mathcal{K}_1,:)$, let $\bar{\mathcal{K}}_1$ and $\bar{\mathcal{K}}_2$ be complement of $\mathcal{K}_1$ and $\mathcal{K}_2$ respectively, then we have,
\begin{eqnarray*}
&M(\mathcal{K}_1,\mathcal{K}_2)=M(\mathcal{K}_1,\mathcal{K}_2)UM(\mathcal{K}_1,\mathcal{K}_2),\\ &M(\mathcal{K}_1,\bar{\mathcal{K}}_2)=M(\mathcal{K}_1,\mathcal{K}_2)UM(\mathcal{K}_1,\bar{\mathcal{K}}_2);\\
& M(\bar{\mathcal{K}}_1,\mathcal{K}_2)=M(\bar{\mathcal{K}}_1,\mathcal{K}_2)UM(\mathcal{K}_1,\mathcal{K}_2) ,\\
& M(\bar{\mathcal{K}}_1,\bar{\mathcal{K}}_2)=M(\bar{\mathcal{K}}_1,\mathcal{K}_2)UM(\mathcal{K}_1,\bar{\mathcal{K}}_2).
\end{eqnarray*}
i.e.,
\begin{eqnarray*}
& M(\mathcal{K}_1,\mathcal{K}_2)=S,\quad M(\mathcal{K}_1,\bar{\mathcal{K}}_2)=SH;\\
& M(\bar{\mathcal{K}}_1,\mathcal{K}_2)=WS,\quad M(\bar{\mathcal{K}}_1,\bar{\mathcal{K}}_2)=WSH,
\end{eqnarray*}
where $W= M(\bar{\mathcal{K}}_1,\mathcal{K}_2)U$, $H=UM(\mathcal{K}_1,\bar{\mathcal{K}}_2)$. 
Hence the results follow.
\end{proof}

We remark that a nonnegative $(r_1,r_2)$-CUR matrix $M$  
is not always a minimal co-$(r_1,r_2)$-separable matrix. In the following, we show an example.
\begin{example} 
\begin{eqnarray*}
M&=&\left(
          \begin{array}{cccc}
                 1  & 1&2&2\\
                 0 & 1&1&2 \\
                 0&0&1&3\\
                 1&2&2&3\\
              \end{array}
            \right)\\
            &=& \left(
          \begin{array}{ccc}
                 1  & 1&2\\
                 0 & 1&1 \\
                 0&0&1\\
                 1&2&2\\
              \end{array}
            \right)\left(
          \begin{array}{ccc}
                 1  & -1&-1\\
                 0 & 1&-1 \\
                 0&0&1\\
              \end{array}
            \right)\left(
          \begin{array}{cccc}
                 1  & 1&2&2\\
                 0 & 1&1 &2\\
                 0&0&1&1\\
              \end{array}
            \right) \\
           & =&M(:,1:3)UM(1:3,:).
\end{eqnarray*}
Hence, $M$ admits $(3,3)$-CUR decomposition, but   
not a co-$(3,3)$-separable matrix.
 \end{example}

\section{Optimization Models and Algorithms}

From Property \ref{Pro:idealmodel}, the 
model for minimal co-$(r_1,r_2)$-separable factorization is given in 
\eqref{eq:idealmodel}. However, in real world, due to the presence of noise, the model \eqref{eq:idealmodel} can be modified to
\begin{equation}\label{eq:rosepmodel}
\begin{split}
\min_{X\in \mathbb{R}^{m\times m}_+, Y\in \mathbb{R}^{n\times n}_+}  \|X\|_{col,0}+\|Y\|_{row,0}, \quad
 \mbox{s.t.} \|M-XM\|\leq \epsilon,\quad \|M-MY\|\leq \epsilon. \nonumber
\end{split}
\end{equation}
where $\epsilon$ denotes the noise level. The norm of $\|M-MX\|$ and $\| M-YM\|$ can be chosen according to the noise level. 
In this paper, we consider the Frobenius norm.

\subsection{Convex Optimization Model and Algorithms}

We observe that optimization problem (\ref{eq:rosepmodel}) can be divided into the following two sub problems and solved independently, that is,
\begin{eqnarray}
\min_{X\in \mathbb{R}^{m\times m}_+}\|X\|_{col,0} ~~ &\mbox{s.t.}&~~ \|M-XM\|\leq \epsilon, \label{eq:sepmodel_X} \\
\min_{Y\in \mathbb{R}^{n\times n}_+}\|Y\|_{row,0}, ~~&\mbox{s.t.}& ~~ \|M-MY\|\leq \epsilon. \label{eq:sepmodel}
\end{eqnarray}
These two sub problems can actually be reduced to the same problem. In the following we only discuss the problem  (\ref{eq:sepmodel}) for simplicity.
Note that  problem (\ref{eq:sepmodel}) is quite challenging to solve, but has been well discussed in reference \cite{gillis2018fast}. Therefore, we will briefly review a fast gradient method presented in this reference. 
According to the reference \cite{gillis2018fast}, problem (\ref{eq:sepmodel}) can be  relaxed to the following convex optimization model:
\begin{equation}\label{eq:convmodel}
\begin{split}
\min_{Y\in \mathbb{R}^{n\times n}_+}trace(Y)  \quad
 \mbox{s.t.}\quad \|M-MY\|\leq \epsilon, 
 0\leq Y(t,l)\leq Y(t,t)\leq 1, 1\leq t,l\leq n.
\end{split}
\end{equation} 
 To avoid column normalization of input matrix $M$,   the 
 optimization problem in (\ref{eq:convmodel}) can be further generalized to the following model for nonscaled matrix.
\begin{equation}\label{eq:nonscamodel}
\begin{split}
 \min_{Y\in \Omega}~~ trace(Y)\quad
 \mbox{s.t.}\quad \|M-MY\|\leq \epsilon,
\end{split}
\end{equation}
where the set $\Omega$ is defined as
\begin{eqnarray*}
\Omega = \{Y\in \mathbb{R}^{n\times n}_+| Y\leq 1, \omega_t Y(t,l)\leq  \omega_l Y(t,t),  1 \leq t,l\leq n \}.
\end{eqnarray*}
where  $\omega_t=\|M(:,t)\|_1$  for all $t$. 

Note that the problem~\eqref{eq:nonscamodel} is smooth and convex.
One may consider
interior-point methods such as SDPT3~\cite{toh1999sdpt3} for solving the problem.
Since this problem contains $ m^2$ variables and many constraints, it would be very expensive to use second order methods. 
In fact, the main aim is to 
identify the important columns 
of $M$ which correspond to the largest entries in the diagonal entries $Y$.
We consider to employ
Nesterov's optimal first-order method~\cite{nes83, nes04} which attains
the best possible convergence rate of $\mathcal{O}(1/k^2)$.
To do so, the following penalized version is considered:
 \begin{align}
 &\min_{Y\in \Omega}F(Y)=\frac{1}{2}\|M-MY\|^2_F  + \lambda \tr(Y),
\label{mod1}
 \end{align}
where $\lambda > 0$ is a penalty parameter which balances the importance between the approximation error $ \frac{1}{2}\|M-MY\|^2_F $ and the trace of  $Y$. The fast gradient method for solving \eqref{mod1} is presented in Algorithm 1. 
We remark in Algorithm 1, the column set $\mathcal{K}$ is identified by a post processing procedure, which is presented as follows:
\begin{itemize}
\item For synthetic data sets, simply pick the $r$ largest entries of the diagonals of Y .
\item For real data sets,  the strategy used in \cite{pan2019generalized} can be adopted, i.e., applying SPA  on $Y^T$ to sort the columns of $M$. Then choose $r$ columns from $M$ according to their sort order. 
\end{itemize}

\begin{algorithm}[h]
\caption{Separable-NMF with a Fast Gradient Method (FGM-SNMF) \cite{gillis2018fast} \label{algo:fgm}}
\begin{algorithmic}[1]
\REQUIRE $M\in\mathbb{R}_{+}^{m\times n}$,
number $r$ of columns to extract,
and
maximum number of iterations $maxiter$.
\ENSURE Matrix  $Y$ solving~\eqref{mod1},
and a set $\mathcal{K} $ of column indices.

\STATE \emph{\% Initialization }

\STATE $\alpha_0 \leftarrow 0.05$;
$L\leftarrow \sigma^2_{\max}(M)$; \\
Initialize  $Y$ and $\lambda$;


\FOR{$k$ = 1 : maxiter}

\STATE \emph{\% Keep previous iterates in memory}

\STATE   $Y_p\leftarrow Y$;

\STATE \emph{\% Gradient computation}

\STATE  $\nabla_Y F(Y)\leftarrow M^TMY-M^TM+\lambda I_n$;

\STATE \emph{\% Gradient step and projection}

\STATE $Y_n \leftarrow \mathrm{P}_\Omega(Y-\frac{1}{L}\nabla_Y F(X,Y))$;

\STATE \emph{\% Acceleration / Momentum step}

\STATE $Y\leftarrow Y_n +\beta_k(Y_n-Y_p)$; \\ where $\beta_k=\frac{\alpha_{k-1}(1-\alpha_{k-1})}{\alpha^2_{k-1}+\alpha_k}$ such that $\alpha_k\geq 0$ and $\alpha^2_k=(1-\alpha_k)\alpha^2_{k-1}$.

\ENDFOR

\STATE $\mathcal{K}\leftarrow \text{post-process}(Y,r)$.
\end{algorithmic}
\end{algorithm}

Therefore, given a matrix $M$, in order to get the row set $\mathcal{K}_1$ and column set $\mathcal{K}_2$, one could directly use Algorithm 1 on $M^T$ and $M$ respectively. However, it is not practical for real applications.  For example, in document classification, the input document-term matrix $M$ could be very sparse. If we identify the important rows $\mathcal{K}_1$ and columns $\mathcal{K}_2$ independently, from Definition \ref{def1}, the core $S=M(\mathcal{K}_1,\mathcal{K}_2)$ may contain some zero columns or rows, which will lead  the loss of some important information. Based on this concern, we propose an alternating fast gradient method for solving CoS-NMF problem.

\subsubsection{Alternating Fast Gradient Method for CoS-NMF}

To prevent to obtain the degenerate of the core matrix $S$, we will utilize the results from Remark \ref{remark:1} that $M(:,\mathcal{K}_2)^T$ is $r_1$-separable, and $M(\mathcal{K}_1, :)$ is $r_2$-separable. More precisely, we consider the following non-scaled convex optimization model derived from \eqref{eq:nonscamodel}.
\begin{eqnarray}
\min\limits_{ X\in \Omega_1} \tr(X) \quad \mbox{s.t.} \quad M_Y=XM_Y, \label{eq:X}\\
\min\limits_{ Y\in \Omega_2} \tr(Y) \quad \mbox{s.t.} \quad M_X=M_XY, \label{eq:Y}
\end{eqnarray}
where $M_Y=M(:,\mathcal{K}_2)$, $M_X=M(\mathcal{K}_1,:)$, and
\begin{eqnarray*}
\Omega_1 &=& \{X\in \mathbb{R}^{m\times m}_+| X\leq 1,  \hat{\omega}_i X(i,j)\leq \hat{\omega}_j  X(j,j), \forall i,j\},\\
\Omega_2 &=& \{Y\in \mathbb{R}^{n\times n}_+| Y\leq 1, \omega_t Y(t,l)\leq \omega_l Y(t,t), \forall t,l\},
\end{eqnarray*}
with $\hat{\omega}_t=\|M_Y(i,:)\|_1$ for all $i$; $\omega_t=\|M_X(:,t)\|_1$  for all $t$.
Note that the row set $\mathcal{K}_1$ is identified by applying post processing on $X$ of \eqref{eq:X}, and the column set $\mathcal{K}_2$ is identified by using post processing on $Y$ of \eqref{eq:Y}, thus we will solve \eqref{eq:X} and \eqref{eq:Y} alternately by using fast gradient method on the following penalized version. 
\begin{equation*}\label{new_op}
\begin{split}
 \min_{X\in \Omega_1}  \dfrac{1}{2}\|M_Y-XM_Y\|^2_F+\lambda trace(X); \\
 \min_{Y\in \Omega_2}   \dfrac{1}{2}\|M_X-M_XY\|^2_F +\lambda trace(Y)
\end{split}
\end{equation*}
This alternating fast gradient method is presented in Algorithm 2, which is referred to as CoS-FGM.

\begin{algorithm}[H]
\caption{Alternating Fast Gradient Method for CoS-NMF \label{algo:alfgm}}
\begin{algorithmic}[1]
\REQUIRE $M\in\mathbb{R}_{+}^{m\times n}$,
number $r_1$ of columns and $r_2$ of rows to extract,
maximum number of iterations $maxiter$, stopping criterion $\delta$.
\ENSURE A set $\mathcal{K}_1$ of column indices and a set $\mathcal{K}_2$ of row indices.

\STATE  $M_X = M$, $M_Y = M$;



\FOR{$k$ = 1 : maxiter}

\STATE \emph{\% Keep previous iterates in memory}

\STATE $M_{XP}\leftarrow M_X$; $M_{YP}\leftarrow M_Y$;

\STATE \emph{\% Update $M_X$} 

\STATE $\mathcal{K}_1=FGM-SNMF(M^T_Y, r_1)$;  \\ $M_X=M(\mathcal{K}_1,:)$;

\STATE \emph{\% Update $M_Y$}

\STATE $\mathcal{K}_2=FGM-SNMF(M_X, r_2)$;\\$M_Y=M(:,\mathcal{K}_2)$;

\STATE \emph{\% Stopping criterion}
\STATE $e =\|M_{XP}-M_{X}\|_F+\|M_{YP}-M_{Y}\|_F$;
\IF{$e  \leq \delta$ }
\STATE break
\ENDIF
\STATE $k=k+1$;
\ENDFOR
\end{algorithmic}
\end{algorithm}
%
%
%
%

\subsection{Factor Matrices $P_1$ and $P_2$}

After both the row set $\mathcal{K}_1$ and column set $\mathcal{K}_2$ are identified by alternating fast gradient method, the core matrix $S$ is then determined by $S=M(\mathcal{K}_1,\mathcal{K}_2)$. From Definition \ref{def1}, the remaining problem is computing factor matrices $P_1$ and $P_2$ by solving the following optimization problem:  given $M\in \mathbb{R}^{m\times n}_+$, $S\in \mathbb{R}^{r_1\times r_2}_+$, find $P_1\in \mathbb{R}^{m\times r_1}_+$ and $P_1\in \mathbb{R}^{r_2\times n}_+$, such that
\begin{equation}
(P_1,P_2)=\mathop{\arg\min}_{U\in \mathbb{R}^{m\times r_1}_+,V\in \mathbb{R}^{r_2\times n}_+} \|M-USV\|^2_F
\end{equation}
We note that this optimization problem is a variant of standard NMF problem. When one of the factors, $U$ or $V$ is fixed, it will be reduced to a convex nonnegative least squares problem (NNLS). We simply use the coordinate descent implemented in \cite{gillis2012accelerated}. The detailed procedure is shown in Algorithm 3.

\begin{algorithm}[h]
\caption{Compute $P_1$ and $P_2$ \label{algo:alsP}}
\begin{algorithmic}[1]
\REQUIRE $M\in\mathbb{R}_{+}^{m\times n}$,
 core matrix $S\in\mathbb{R}_{+}^{r_1\times r_2}$,
maximum number of iterations $maxiter$, stopping criterion $\delta$.
\ENSURE $P_1$ and $P_2$.
 
\STATE $M_1=M(\mathcal{K}_1,:)$; $M_2=M(:,\mathcal{K}_2)$.
\STATE  \emph{\% Generate initial $P_1$ and $P_2$}

\STATE  $P_1= \mathop{\arg\min}\limits_{U\in \mathbb{R}^{m\times r_1}_+}\|M-UM_1\|^2_F$;\\
 $P_2= \mathop{\arg\min}\limits_{V\in \mathbb{R}^{r_2\times n}_+}\|M-M_2V\|^2_F$;

\FOR{$k$ = 1 : maxiter}

\STATE \emph{\% Keep previous iterates in memory}

\STATE $\hat{P}_{1}\leftarrow P_1$; $\hat{P}_{2}\leftarrow P_2$;


\STATE $W=P_1S$; \quad
solve $P_2$ from $\min\limits_{V\in \mathbb{R}^{r_2\times n}_+}\|M-WV\|^2_F$;

\STATE  $H= SP_2$; \quad
solve $P_1$ from $\min\limits_{U\in \mathbb{R}^{m\times r_1}_+}\|M-UH\|^2_F$;

\STATE \emph{\% Stopping criterion}
\STATE $e =\|\hat{P}_{1}-P_{1}\|_F+\|\hat{P}_{2}-P_2\|_F$;
\IF{$e  \leq \delta$ }
\STATE break
\ENDIF
\STATE $k=k+1$;
\ENDFOR
\end{algorithmic}
\end{algorithm}

%
%
%
%

\section{Numerical Experiments}

In this section, we show the performances of the proposed CoS-NMF model on synthetic datasets (Section~\ref{sec:synth}),  document datasets (Section~\ref{sec:doc}) and facial database (Section~\ref{sec:facial}). All experiments were run on Intel(R) Core(TM) i5-5200 CPU @2.20GHZ with 8GB of RAM using Matlab.

We compared our model with the several  state-of-the-art methods. The compared algorithms are briefly summarized as follows,
\begin{enumerate}

\item SPA (Successive projection algorithm\cite{araujo2001successive,gillis2014fast,fu2015self}) is a state-of-the-art separable NMF method. It selects the column with the largest $l_2$ norm and projects all columns of $M$ on the orthogonal complement of the extracted column at each step. Note that SPA can only identify a subset of the columns of the input matrix M, therefore we consider the following three variants.

\begin{itemize}
\item  SPA$^+$: We apply SPA  on $M^T$ to identify $r_1$ important rows of $M$, and then on $M$ to identify $r_2$ important columns of $M$. 

\item SPAR: We apply SPA on $M^T$ to identify $r_1$ rows.

\item SPAC: We apply SPA on $M$ to identify $r_2$ columns.
\end{itemize}

%
%

\item GSPA (Generalized SPA, \cite{pan2019generalized}) is a state of the art generalized separable NMF algorithm. It is a fast heuristic algorithm driven from SPA, applied on $M$ to identify $r$ columns and rows of $M$. 

\item GS-FGM (Generalized separable fast gradient method,\cite{pan2019generalized}) is an other state of the art generalized separable NMF algorithm which is based on fast gradient method on separable NMF.

\item BiOR-NM3F (Bi-orthogonal tri-factorization \cite{ding2006orthogonal}) is an algorithm for co-clustering, aims to solve problem (\ref{trinmf}).

\item A-HALS algorithm is a state-of-the-art NMF algorithm, namely the accelerated hierarchical alternating least squares algorithm~\cite{gillis2012accelerated}.

\item MV-NMF  is a state-of-the-art minimum-volume NMF algorithm~\cite{fu2016robust} which uses a fast gradient method to solve the sub problems in $W$ and $H$ from~\cite{leplat2019minimum}. 
\end{enumerate}

For simplicity, the methods that select important columns and rows from the input matrix, i.e., SPA, GSPA, GS-FGM and the proposed CoS-FGM, are referred to as column-row selected methods.

\begin{remark}
We  have also tested other separable NMF algorithms: successive nonnegative projection algorithm (SNPA \cite{gillis2014successive}) and XRAY  \cite{kumar2013fast}. They showed similar results as SPA. We also applied fast gradient method (FGM\cite{gillis2018fast}) to identify $r_1$ rows and $r_2$ columns of $M$, and found that its results are not as good as CoS-FGM and SPA. Hence for simplicity, we do not present their results here. 
\end{remark}


The stopping criterion of  CoS-FGM and GS-FGM :
We will use $\delta = 10^{-6}$ for synthetic data sets and $\delta = 10^{-2}$ for the real data sets (document data sets and facial database). The maximum iteration is set to be 1000. For BiOR-NM3F and both NMF algorithms, we use the default parameters and perform 1000 iterations.

\subsection{Synthetic data sets} \label{sec:synth}


In this section, we compare the algorithms on synthetic data set generated fully randomly. We identify the subsets $\mathcal{K}_1$ and $\mathcal{K}_2$ by using the $r_1$ largest diagonal entries of $X$ and $r_2$ largest diagonal entries of $Y$, respectively.

Given the subsets
$(\mathcal{K}_1,\mathcal{K}_2)$ computed by an algorithm, in order to show the effect of these algorithms, we will report the following two quality measures:
\begin{enumerate}

  \item The accuracy is defined as the proportion of correctly identified row and column indices:
\begin{equation} \label{accuracy}
  \text{accuracy} =
  \frac{|\mathcal{K}^*_1\cap\mathcal{K}_1| + |\mathcal{K}^*_2\cap\mathcal{K}_2|}{|\mathcal{K}^*_1| + |\mathcal{K}^*_2|},
\end{equation}
where $\mathcal{K}^*_1$ and $\mathcal{K}^*_2$ are the true   row and column indices used to generate $M^*$. 
 
Note that  BiOR-NM3F and both NMF algorithms  do not identify columns and rows from the input matrix, hence,  
the accuracy cannot be computed.

 \item Since the models to represent data matrix are different. To be fair, the relative approximation  for CoS-FGM and SPA$^+$ is defined as
 \begin{equation} \label{appro1}
 1- \frac{\min_{P_1 \geq 0, P_2 \geq0}\|M-P_1M(\mathcal{K}_1, \mathcal{K}_2) P_2 \|_F}{\|M\|_F}.
\end{equation}
For GSPA and GS-FGM, the relative approximation  is defined as
 \begin{equation}  \label{appro2}
 1- \frac{\min_{P_1 \geq 0, P_2 \geq0}\|M-P_1M(\mathcal{K}_1, :)-M(:,\mathcal{K}_2) P_2 \|_F}{\|M\|_F}.
\end{equation}

For the rest methods, we compute the relative approximation as 
$1- \frac{\min \|M-\tilde{M} \|_F}{\|M\|_F}$, where $\tilde{M}$ is the approximation of $M$  obtained by these methods. 
\end{enumerate}


\subsubsection{Fully randomly generated data} \label{sec:randnoise}

We generate noisy co-$(10,3)$-separable matrices $M \in \mathbb{R}^{100\times 100}$ as follows:
$$
\Pi_r \max
\left( 0 ,
 \underbrace{  D_r
\left(
     \begin{array}{cc}
       S & SH \\
       WS & WSH \\
     \end{array}
   \right)  D_c
    }_{M^s}
   + N \right) \Pi_c.
$$
Here we consider the following settings.
%

  \noindent $\bullet$ The entries of the matrices
 $S\in \mathbb{R}^{10\times 3}$, $W \in \mathbb{R}^{90 \times 10}$  and   $H \in \mathbb{R}^{3 \times 97}$ are generated uniformly at random in the interval [0,1] by the \texttt{rand} function of MATLAB. 

   \noindent $\bullet$ The diagonal matrices $D_r$ and $D_c$ are computed by the algorithm in \cite{knight2008sinkhorn, olshen2010successive} that alternatively scales the columns and rows of the input matrix, such that 
 $M^s$ is scaled.

 \noindent $\bullet$ The noise matrix $N \in \mathbb{R}^{100 \times 100}$ is generated  at random normal distribution by the \texttt{randn} function of MATLAB.  We normalize $N$ such that $||N||_F = \epsilon ||M^s||_F$,
where $M^s$ is the noiseless scaled co-$(10,3)$-separable matrix, and $\epsilon$ is a parameter that relates to the noise level.

  \noindent $\bullet$ $\Pi_r$ and $\Pi_c$ are  permutation matrices generated randomly.

In this experiment, GS-FGM is run with the parameter $\tilde{\lambda} = 0.25$. We note that $M^s$ is co-$(10,3)$ separable matrix, its nonnegative rank is not larger than $3$, fairly, the value of factorization rank for both NMF algorithms is hence set to be $r=3$. For GSPA and GS-FGM, we set $(r_1,r_2)=(10,3)$ to test the accuracy of identified row and columns, even though it is not fair to compare to their relative approximation since the factorization rank of their generalized separable representation is $r_1+r_2=13$.

\begin{figure}[h]
\includegraphics[width=\textwidth]{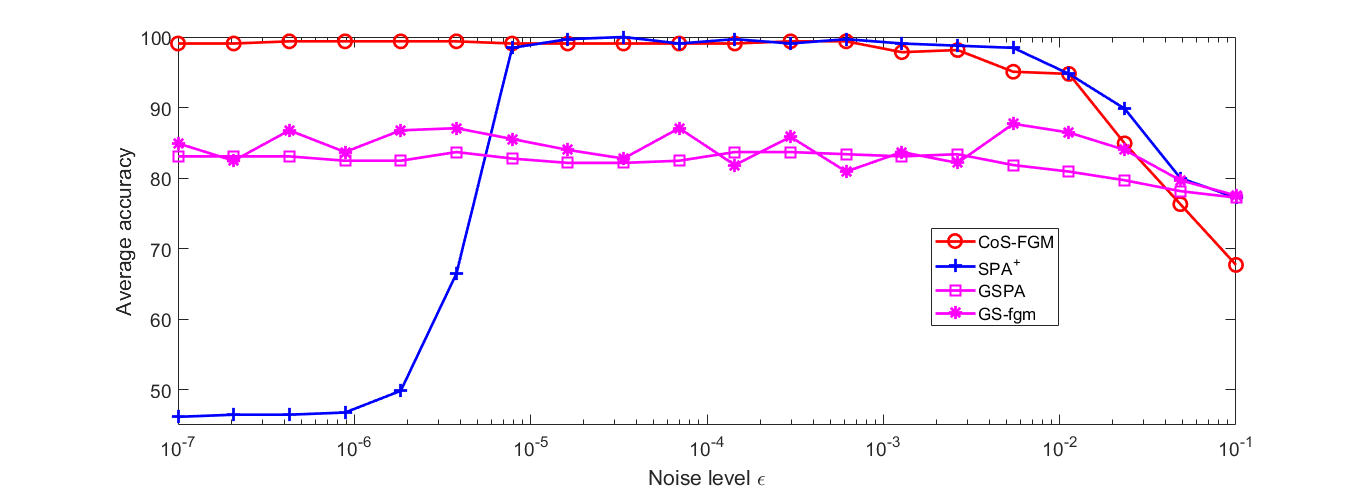}
\caption{Average accuracy~\eqref{accuracy} in percentage for the different algorithms on the fully randomly generated CoS-NMF matrices.  \label{fullacc}}
\end{figure}

\begin{figure}[h]
\includegraphics[width=\textwidth]{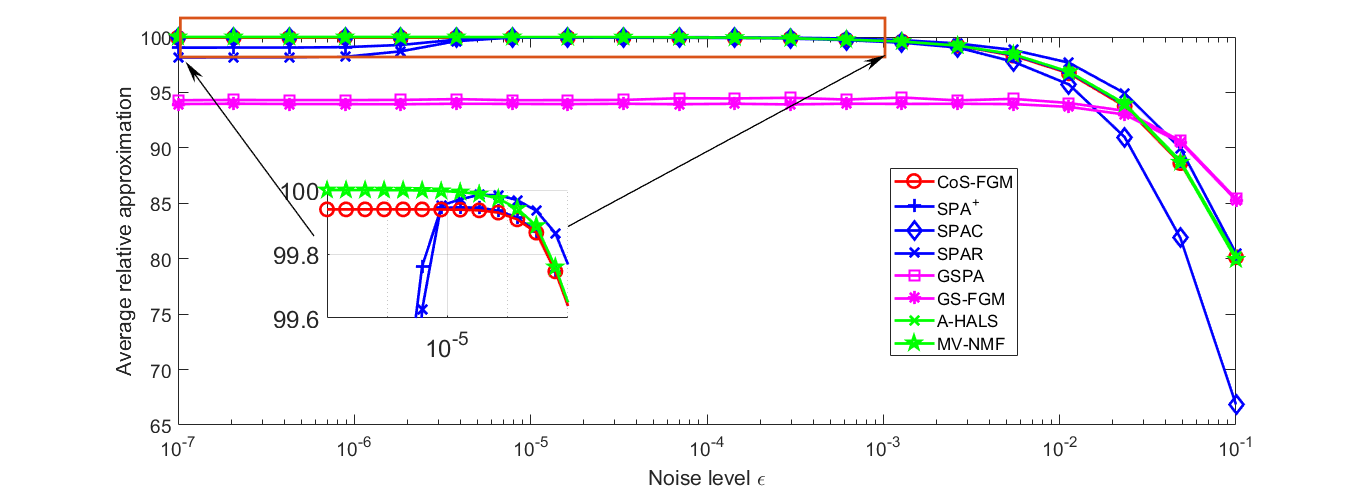}
\caption{Average relative approximation~\eqref{appro1} - \eqref{appro2} in percentage on the fully randomly generated CoS-NMF matrices.
\label{fullres}}
\end{figure}

We use 20 noise levels 
$\epsilon$ logarithmically spaced in $[10^{-7},10^{-1}]$ (in MATLAB, \texttt{logspace(-7,-1,20)}).
For each noise level, we generate 25 such matrices and report the average quality measures in percent on
Figs.\ref{fullacc}-\ref{fullres}. We have the following observations:

 \noindent $\bullet$ In terms of accuracy, CoS-FGM has an accuracy of nearly  $100 \%$ for all $\epsilon \leq 0.0026$. For low noise levels $\epsilon \leq 7.85\times 10^{-6}$, CoS-FGM performs the best, but SPA$^+$  is not able to recover column and row indices. When the noise becomes larger, the accuracy of SPA$^+$ increases. The accuracies of both of CoS-FGM and SPA$^+$ decrease when the noise is larger than $0.0026$. As expected, we find that the performances of GS-FGM and GSPA are worse than CoS-NMF in most noise levels.

 \noindent $\bullet$ In terms of relative approximation, both NMF methods performs similarly as CoS-FGM. For for all $\epsilon \leq 0.0026$,  they can almost  $100 \%$ approximate the input matrix.  This is not surprising since NMF factorizes the input matrix with no other constraints than nonnegativity. It is actually nice  to find that the solutions from CoS-FGM can generate the same approximation with NMF even though CoS-NMF is much more constrained. The reason is that the input data satisfies the CoS-NMF assumptions.
 
\begin{remark}
We have tested BiOR-NM3F and found that its relative approximations for all noise level $\epsilon$ are smaller than  $20 \%$, hence we do not present its results in Fig~\ref{fullres}.
\end{remark}

%
%
%
%
%
%

\subsection{Document Data Sets} \label{sec:doc}

In this section, we test these  methods on document data sets including TDT30 data set~\cite{cai2008modeling}, 
and the 14 data sets from~\cite{zhong2005generative}. Note that for Newsgroups 20, which is a very large data set, we only consider the first 10 classes and refer to the corresponding data set as NG10. Here we do not scale the input matrix since these document data sets are very sparse. We will consider the clustering abilities of the methods on both  words and document terms, however, for all the document datasets, only the clustering ground truth on document terms are provided. Hence to determine the clustering ground truth on words, we will first pre-process the datasets. 

\textbf{Preprocessing} Given a document-word data matrix $M_0\in \mathbb{R}^{m\times n}$ , we select the top 1000 words from $M_0$ to construct the new document-word data  matrix $M\in \mathbb{R}^{m\times 1000}$. We then label the word  according to the label of its associated document  in which the appearing probability of this word is the largest. In this way, we get the clustering ground truth on words.

For these new document-word data sets, though the number of words is 1000, the number of documents are still very large. 
For some methods like GS-FGM that requires $\mathcal{O}(mn^2+nm^2)$ computation operations, it is impractical to apply these methods directly on large data matrix.  Hence, we will employ a similar strategy in \cite{gillis2018fast,pan2019generalized}, use the hierarchical clustering \cite{gillis2015hierarchical} that runs in $\mathcal{O}(mnlog_2 C)$ ($C$ is the number of the clusters to generate),  to preselect a subset of columns and
rows from the input matrix. Precisely, for all these data sets except tr11 and tr23, we extract 500 documents and 500 words, and consider a submatrix matrix $M_s \in \mathbb{R}^{500\times 500}$. For tr11 and tr23 data sets, since the number of documents is relatively small (414 for tr11, 204 for tr23), we keep all the documents and extract 500 words. Here we take into account the importance of each selected column and row by identifying the  number of data points attached to it (this is given by the hierarchical clustering). We scale it using the square root of the number of points belonging to its cluster.

We apply the column-row selected methods (CoS-FGM, SPA, GSPA, GS-FGM) on the subsampled matrix to identify a subset of $r_1$ rows and $r_2$ columns. From these subsets, we can then identify their corresponding columns and rows in the original data matrix. 

For all the methods, we will report the clustering accuracy used in \cite{kuang2015symnmf,moutier2021off} that quantifies the level of correspondence between the clusters and the ground truth, defined as 
\begin{equation}\label{Acc}
Acc=1-\max_{\Pi\in [1,2,\cdots,r]}\sqrt{\frac{\|Q_{\Pi}-Q^*\|_F}{rn}}\in [0,1],
\end{equation}
where $[1,2,\cdots,r]$ is the set of permutations of $\{1,2,\cdots, r\}$ and $Q_{\Pi}$ is the clustering matrix $Q$ whose columns are rearranged according to the permutation $\Pi$, $Q^*$ is the clustering ground truth. We will also report the approximation quality measure defined in Section \ref{sec:synth}.

Given document-word data matrix,  the clustering matrix $Q$ is computed in the following way. 
\begin{itemize}
\item For those methods (CoS-FGM, SPA, BiOR-NM3F) that compute $M\approx P_1SP_2$, $Q_1$ and $Q_2$ represent the clustering matrix of document and word respectively, which can be obtained by using hard clustering on $P_1$ and $P_2$, i.e., $Q_{i,j}=1$ if $j=argmax_t\{P(i,t)\}$, and $Q_{i,j}=0$ for else. 

\item For GSPA and GS-FGM that compute $M\approx W_1H_1+W_2H_2= [W_1, W_2][H_1, H_2]^T$, $Q_1$ and $Q_2$ can be obtained by using hard clustering on its factor matrices $[W_1,W_2]$ and $[H_1, H_2]$ respectively. %

\item For the rest methods (SPAR, SPAC, A-HALS and MV-NMF) that compute $M\approx WH$, $Q_1$ and $Q_2$ can be obtained by using hard clustering on its factor matrices $W$ and $H$ respectively.
\end{itemize}

In this experiment, for GS-FGM, we try 10 different values of $\lambda$ from $[10^{-3}, 10]$ with 10 log-spaced values (in MATLAB, logspace(-3,1,10)), and keep the solution with the highest approximation quality. Since the cluster number of words is equal to that of documents, we then set $r_1=r_2=r$ for all document datasets. The results are presented in Table \ref{docus_approx}, \ref{docus_cluster} and \ref{word_cluster}.
Here we have the following observations.

(i) In terms of approximation ability, among all the column-row selected methods,  CoS-FGM gets the highest in 10 out of the 15 datasets, and has the highest average approximation. We note that the average approximation of CoS-FGM is only a bit less than that of NMF methods.   This is actually very promising because the NMF is much less constrained compared to CoS-NMF model.  

(ii) In terms of the clustering ability, CoS-FGM has the highest average accuracies in both document and word clustering among all the methods.

(iii) The last line of Table \ref{docus_approx}  reports the average computational time in seconds for these algorithms. CoS-FGM is slower but the computational time is reasonable since it needs to run fast gradient method iteratively. Among all the methods, BiOR-NM3F takes more than 4000 seconds, is the slowest, and all SPA variants are the fastest, take not more than 0.01 seconds.

\begin{table*}[h]
  \centering
  \resizebox{\textwidth}{30mm}{
  \begin{tabular}{|c||c|c||ccc||cccc||c||cc|}
    \hline
    Dataset&r&CoS-FGM&SPA$^+$&SPAC&SPAR&$r$&GSPA&$r$&GS-FGM&BiOR-NM3F&A-HALS&MV-NMF\\
    \hline
    NG10&10& 93.83&93.85 & 93.48&93.62&(9,1)&93.76&(9,1)&93.76 &0.01&94.31&94.24\\

   TDT30&30& 24.64  &22.03 &20.91&17.96&(7,23)&20.98&(10,20)&21.20 &3.94&26.03&25.93\\

   classic&4& 6.66 &6.66 & 2.89& 2.56 &(1,3)&2.63&(3,1)&2.93&1.36&7.06&5.96\\
   
   reviews&5& 17.54 &14.73 &15.01&10.64 &(2,3)&15.07&(2,3)&15.07 &1.49&19.31&19.24\\
   
   sports&7& 16.37 &16.15 &13.53&10.62 &(0,7)&13.53&(0,7)&13.53 &5.71&17.51&17.40\\
   
    ohscal&10& 15.51 &14.50 & 13.62&11.41 &(0,10)&13.62&(0,10)&13.62 &5.58&15.84 &15.85\\
    
    k1b&6&13.45& 12.05 & 10.20&7.79 &(2,4)&9.83&(2,4)&9.83 &3.09&13.80&13.65\\
    
    la12&6&11.68&11.33 &7.35&5.50 &(3,3)&5.82&(1,5)&6.81 &3.51&11.85&11.56\\
    
    hitech&6&11.31&11.08 & 8.02&6.48 &(3,3)&8.62&(3,3)&9.27 &3.93&12.47&12.35\\
    
    la1&6& 11.85& 10.45&6.88&6.53 &(0,6)&6.88&(1,5)&7.64 &3.61&12.00&11.66\\
    
    la2&6&12.05& 11.39 & 8.50& 7.02 &(1,5)&8.47&(1,5)&8.47 &3.16&12.18&11.99\\
    
    tr41&10&56.76& 59.15 &58.32&60.23 &(8,2)&60.43&(9,1)&53.94 &15.02&61.23&60.78\\
    
    tr45&10& 73.26&73.15 &71.25&75.15 &(10,0)&75.15&(10,0)&75.15 &29.38&78.22&78.18\\
    
    tr11&9& 73.98& 75.01 & 65.44 &76.35 &(6,3)&76.10&(6,3)&76.10 &13.92&78.46&78.41\\
    
    tr23&6&71.68& 70.99 &67.44&71.74~ &(2,4)&66.61&(4,2)&71.47 &8.77&73.37&73.33\\
    \hline
    average&--&34.04 & 33.50 & 30.86 &  30.91  & --  & 31.83  & -- &31.92& 6.83&  35.58& 35.37 \\
    \hline
    time &-- &36.27s&0.01s  &0.004s & 0.005s  &  --  & 0.05s & -- & 0.41s&4174.8&28.32s& 170.27s \\
    \hline
   \end{tabular}}
  \caption{The relative approximation quality \eqref{appro1} - \eqref{appro2} in percentage  for the document data sets.The last line reports the average computational time in seconds for the different algorithms.  \label{docus_approx} }
\end{table*}

\begin{table*}[h]
  \centering
  \resizebox{\textwidth}{30mm}{
  \begin{tabular}{|c||c|c||ccc||cccc||c||cc|}
    \hline
    Dataset&r&CoS-FGM&SPA$^+$&SPAC&SPAR&$r$&GSPA&$r$&GS-FGM&BiOR-NM3F&A-HALS&MV-NMF\\
    \hline
    NG10&10& 61.59&61.48& 59.04&60.42&(9,1)&60.41&(9,1)&60.42 &59.21&61.03&60.36\\

   TDT30&30& 83.40&79.70 &79.42&81.33 &(7,23)&79.06&(10,20)&80.26 &76.74&80.49&80.99\\

   classic&4& 51.74&52.22 & 41.49&44.57 &(1,3)&42.23&(3,1)&46.02 &43.55&52.49&51.15\\
   
   reviews&5& 56.21&60.49 & 49.58&60.40 &(2,3)&54.63&(2,3)&54.71 &50.00&56.50&58.55\\
   
   sports&7& 63.49&62.09 & 55.44&60.96 &(0,7)&55.49&(0,7)&55.36 &54.57&59.38&58.62\\
   
    ohscal&10& 64.63&63.82 &62.06 &62.67 &(0,10)&62.01&(0,10)&62.18 &60.32&62.46&61.90\\
    
    k1b&6& 69.45&62.01 &54.36& 57.55 &(2,4)&55.15&(2,4)&55.15 &59.84&63.56&64.67\\
    
    la12&6&60.28&56.69 &51.46&55.78 &(3,3)&53.98&(1,5)&53.82 &52.30&55.38&54.60\\
    
    hitech&6& 57.98&56.02 &55.40&55.14 &(3,3)&57.70&(3,3)&57.96 &53.35&58.24&58.60\\
    
    la1&6& 55.94&59.79 &56.30&52.73&(0,6)&55.96&(1,5)&56.98 &52.82&59.23&56.45\\
    
    la2&6&55.40&51.93 & 52.50&54.58 &(1,5)&53.11&(1,5)&53.09 &53.19&54.55&53.69\\
    
    tr41&10&62.88&67.81 & 65.12&65.95 &(8,2)&65.75&(9,1)&66.93 &67.88&66.56&64.52\\
    
    tr45&10& 63.96&63.45& 64.70&64.21 &(10,0)&64.21&(10,0)&64.21 &63.33&64.13&64.86\\
    
    tr11&9& 62.43&63.59 &66.67&62.07&(6,3)&66.91&(6,3)&66.91 &61.44&67.24&67.32\\
    
    tr23&6& 55.72&54.45 & 52.17&56.28&(2,4)&55.17&(4,2)&55.17 &54.27&55.35&54.27\\
    \hline
    average & --&\textbf{61.67}&61.04&57.71&59.64&--&58.78&--&59.28&57.52&61.11&60.70
    \\
    \hline
   \end{tabular}}
  \caption{The accuracy of documents clustering  \eqref{Acc} in percentage  for the document data sets.\label{docus_cluster} } 
\end{table*}

\begin{table*}[h]
  \centering
  \resizebox{\textwidth}{30mm}{
  \begin{tabular}{|c||c|c||ccc||cccc||c||cc|}
    \hline
    Dataset&r&CoS-FGM&SPA$^+$ &SPAC&SPAR&$r$&GSPA&$r$&GS-FGM&BiOR-NM3F&A-HALS&MV-NMF\\
    \hline
    NG10&10&64.76&63.48& 65.10&61.06&(9,1)&61.17&(9,1)&61.14 &59.90&62.72&61.89\\

   TDT30&30&79.47&77.73 &77.95&77.74&(7,23)&77.46&(10,20)&77.98 &76.78&79.33&79.44\\

   classic&4& 68.46&60.63 & 49.35&40.00 &(1,3)&51.52&(3,1)&42.60 &40.52&48.77&52.62\\
   
   reviews&5& 60.66&50.81 & 49.72&48.85 &(2,3)&50.85&(2,3)&50.85 &46.31&56.83&55.68\\
   
   sports&7& 60.65& 56.91 & 56.64 &53.93 &(0,7)&56.64&(0,7)&56.64 &51.98&57.00&56.71\\
   
    ohscal&10& 67.97&62.50 & 63.56&59.16 &(0,10)&63.56&(0,10)&63.56 &58.82&65.19&66.15\\
    
    k1b&6& 60.76&56.22 &54.36&49.01 &(2,4)&52.99&(2,4)&53.10 &48.42&57.89&60.88\\
    
    la12&6& 61.75&60.97 & 51.04&51.76 &(3,3)&52.60&(1,5)&54.43 &49.08&62.99&60.63\\
    
    hitech&6&53.81&51.94 &48.78&49.37 &(3,3)&51.11&(3,3)&50.87 &48.68&52.32&50.30\\
    
    la1&6&58.93& 55.24 & 50.07&51.01 &(0,6)&50.07&(1,5)&49.93 &49.17&62.27&60.13\\
    
    la2&6& 58.81&58.25&52.88&49.77 &(1,5)&52.64&(1,5)&52.64 &48.65&57.65&56.87\\
    
    tr41&10&63.72&61.79 &61.95&61.56&(8,2)&61.45&(9,1)&61.40 &61.27&62.16&62.99\\
    
    tr45&10& 64.93& 65.62 & 64.33 &67.72 &(10,0)&67.66&(10,0)&67.84 &64.76&65.36&64.42\\
    
    tr11&9& 64.07& 61.18 & 64.32&60.87 &(6,3)&61.47&(6,3)&61.53 &58.37&64.50&62.73\\
    
    tr23&6& 51.90& 57.85 & 58.65&59.05 &(2,4)&62.14&(4,2)&53.88 &53.67&53.96&54.83\\
     \hline
    average & --&\textbf{62.71}&60.07&57.91&56.06&--&58.22&--&57.23&54.43&60.60&60.42
  \\
    \hline
   \end{tabular}}
  \caption{The accuracy of words clustering \eqref{Acc} in percentage  for the document data sets.\label{word_cluster} }
\end{table*}

\begin{figure}[h]
 \centering
\includegraphics[width=1.05\textwidth]{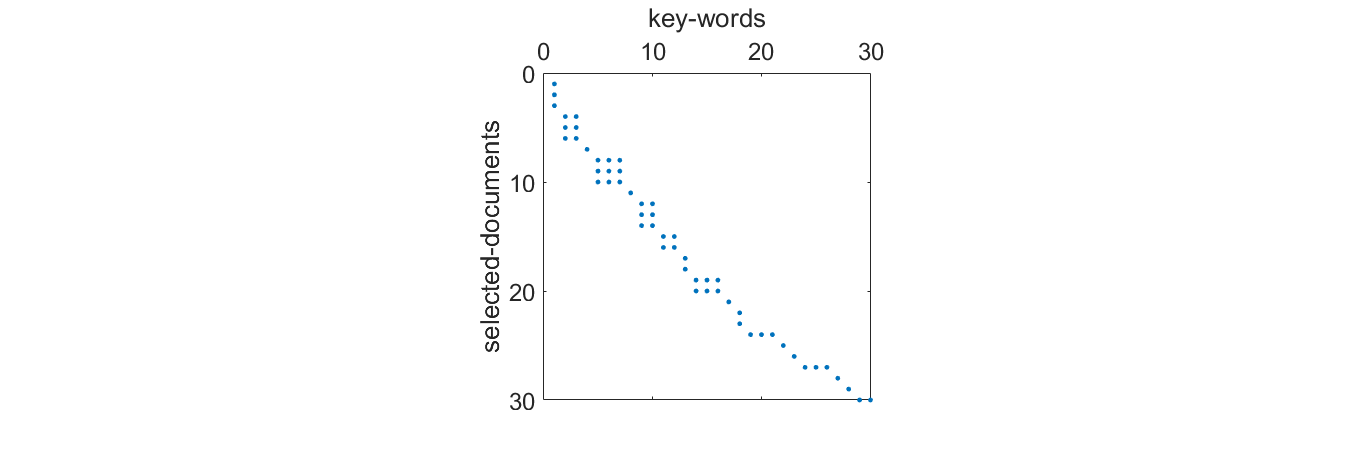}
\caption{ Interpretation of core matrix from CoS-FGM method on TDT30 dataset.  \label{core-doc-word}}
\end{figure}

In particular, we present the key words  in Table \ref{keyword}  and show the interpretation of the core matrix from CoS-FGM method on TDT30 dataset in Fig. \ref{core-doc-word}. Since the ground truths of these 30 selected words and 30 important documents have been given, we cluster the documents and words based on their common topic. For example, there are three documents,  sharing a same key word in the first group (topic); in 4th group (topic), three keys words appear in three important documents.  The results in  Fig. \ref{core-doc-word} verify the assumption of the core matrix of CoS-NMF, i.e., for a topic, there are at least one ''pure'' document contains at least one  anchor word. 

It is interesting to find that these pure documents and key words are only selected from 18 topics, while the TDT30 has 30 topics. We assume that the reason is  these 18 topics are more important than the others. To verify our assumption, we observe that the number of the words belongs to these selected topics is 742, accounts for 74.2$\%$ of total 1000 words, and the number of the documents belongs to these topics is 8333, accounts for 88.71$\%$ of 9394 documents.

\begin{table*}[h]
  \centering
  \begin{tabular}{ c|c||c|c||c|c }
    \hline
    label&words&label&words&label&words\\
    \hline
   1 &  'index' &2&  'tripp',
    'allegations' &3&'death'\\  
    \hline
    4& 'church', 'pope','cuba'&5&'downhill'&6&'saudi',    'cohen'\\
\hline
7& 'super', 'denver'&8&'police'&9&'vote', 'hindu',    'election'\\
\hline
10&'tax'&11&'viagra'& 12&'school', 'voice',    'children'\\
\hline
13& 'tests'&14& 'netanyahu'&15& 'students',    'habibie',   'suharto'\\
\hline
16& 'kaczynski' &17&  'kaczynski'&18& 'bulls',    'jordan'\\
    \hline
    \end{tabular} 
  \caption{The key words from the important documents on TDT30 of CoS-FGM\label{keyword}}
\end{table*}

\subsection{Facial Database}\label{sec:facial}

In this section, we apply the algorithms on facial database: ORL Database of Faces which contains 400 facial images taken at the Olivetti Research Laboratory in Cambridge between 
April 1992 and April 1994. There are 40 distinct subjects, each subject has ten different images and each image is size of 112 x 92. Here, we resize each image to the size of $23\times 19$,  normalize the pixel value to $[0,1]$ and form data vectors of dimension 437.  The "pixel $\times$ image" matrix is size of $437\times 400$.

We used the same way as in document datasets to tune the best parameter $\lambda$ for GS-FGM method. Since the ground truth of facial image clusters are given, i.e., the same subjects are regarded as the same cluster, we can use the same strategy  for document dataset to compute the cluster accuracy of their facial images. Note that we do not have clustering ground truth for pixels, hence, we can choose the number $r_1$ of pixels arbitrarily. Here, we test CoS-FGM and SPA$^+$ by letting $r_1=40$ and $r_1=80$ respectively. $r_2$ is referred to as the cluster number of facial images, i.e., $r_2=40$. For the rest algorithms, the factorization rank is set to be the cluster number of facial images, i.e., $r=40$.  

In Table \ref{table:facial}, we report relative approximation quality \eqref{appro1} - \eqref{appro2} and the cluster accuracy \eqref{Acc} for images 
of these methods. Here we have the following observations. 

(i) In terms of approximation quality, among all the in the column-row selected methods, CoS-NMF has the highest approximation for the case of $r=(80,40)$. BioR-NM3F has the lowest approximation due to the constraints of BioR-NM3F model itself. 

(ii) In terms of clustering accuracy for subjects,  CoS-NMF has the highest accuracy for both cases of  $r=(40,40)$ and $r=(80,40)$. We also notice that both generalized separable NMF methods (GSPA and GS-FGM) do not perform well in this clustering task.

(iii) Note that when the number $r_2$ of selected pixels is increasing, both approximation quality and clustering accuracy increase for CoS-FGM and SPA$^+$.

\begin{table*}[h]
  \centering
   \resizebox{\textwidth}{10mm}{
  \begin{tabular}{|c||ccc||ccc||cccc||c|c|c|cc|}
    \hline
    Dataset&r&CoS-FGM&SPA$^+$&r&CoS-FGM&SPA$^+$&$r$&GSPA&$r$&GS-FGM&$r$&SPAC&BioR-NM3F&A-HALS&MV-NMF\\
    \hline
    Appro &(40,40)&82.22&81.66  &(80,40)&83.08&82.45 &(21,19)&81.51&(10,30)&83.02 &40&82.90&1.36&89.45&87.95\\
    \hline
    Acc&--&84.67&82.64&--&84.83&83.27& --&80.28&--&80.83&--&83.01  &83.04&81.77&83.99\\
    \hline
   time & --&10.63s&0.01s& --&20.31s&0.02s&--&0.30s&--&0.34s&--&0.01s &72.32s&3.42s&71.46s
   \\  
    \hline
   \end{tabular} }
  \caption{The relative approximation quality and clustering accuracy in percentage  for ORL database.  The last line reports the computational time in seconds for the different algorithms. \label{table:facial} }
\end{table*}

\begin{figure}[h]
 \centering
\includegraphics[width=0.2\textwidth]{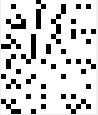}
\caption{The key  pixels to generate core matrix from CoS-FGM.  \label{key-pixels}}
\end{figure}

\begin{figure}[h]
 \centering
\includegraphics[width=\textwidth]{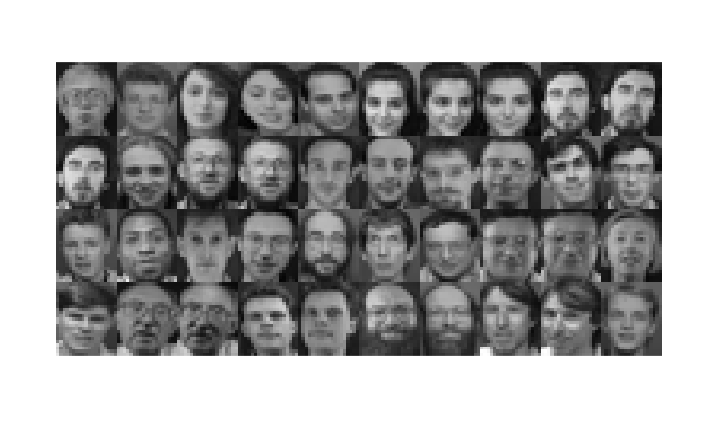}
\caption{The important facial images  to generate core matrix from CoS-FGM.  \label{key-subjects}}
\end{figure}

\begin{figure}[h]
 \centering
\includegraphics[width=\textwidth]{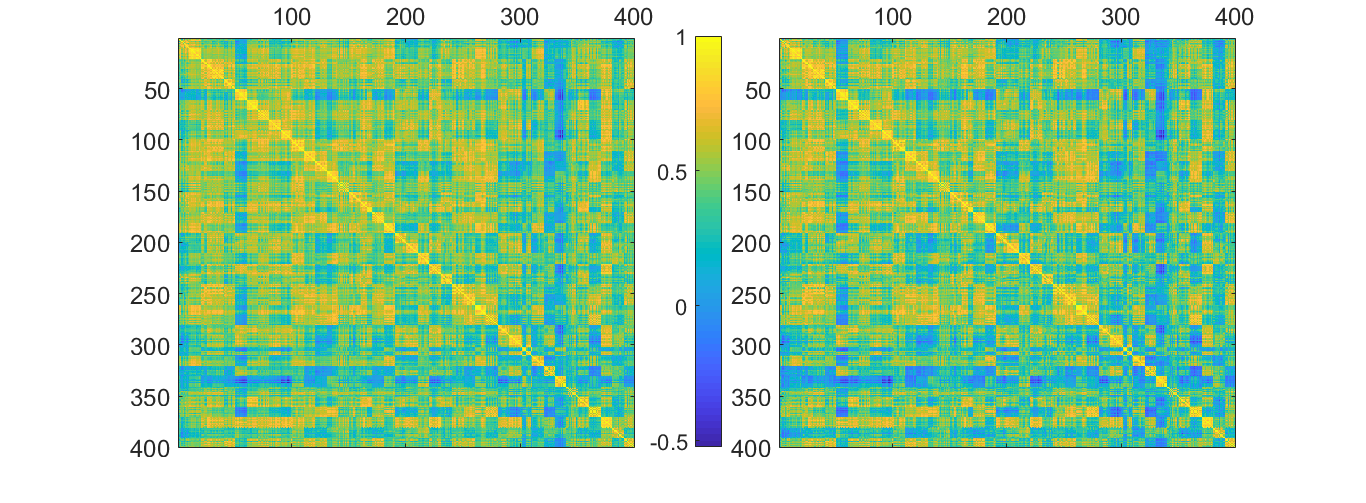}
\caption{From left to right: ground truth, clustering result  on correlation matrix formed by $M(:,\mathcal{K}_2)$,  clustering result  on correlation matrix formed by $M(\mathcal{K}_1,\mathcal{K}_2)$.  \label{correlation}}
\end{figure}

In Figs. \ref{key-pixels}-\ref{key-subjects}, we present the selected 40 important images and 80 key pixels in core matrix from CoS-FGM for the case of $r=(80,40)$.  We remark that the  pixels in Fig.\ref{key-pixels} are selected from the facial images, to test how good these pixels are, we consider the following strategy: Let the index set of the selected pixels be $\mathcal{K}_1$, the facial image matrix with the selected pixels is $M_r=M(\mathcal{K}_1,:)$, where $M$ is the ORL facial  matrix. We compute correlation coefficient matrices of $M\in \mathbb{R}^{437\times 400}$ and  $M_r\in \mathbb{R}^{80\times 400}$ respectively, denoted as $corr(M)$ and $corr(M_r)$.  We present their correlation coefficient results in Fig. \ref{correlation}. It shows that in both figures, the correlation values of facial images belong to same cluster are higher than the others. The relative error denoted as $\dfrac{\|corr(M)-corr(M_r)\|_F}{\|corr(M)\|_F}$ is $0.1663$.  These results are quite encouraging.

\section{Conclusion}

In this paper, we have generalized separability condition to co-separability on NMF problem: instead 
of only selecting columns of the input matrix to approximate
it, we select columns and rows to form a sub-matrix to represent the input matrix.  We refer
to this problem as co-separable NMF (CoS-NMF). We studied some mathematics properties of CoS-NMF matrices that
can be decomposed using CoS-NMF. In particular, we discussed the relationships between CoS-NMF and other related matrix factorization models: CUR decomposition, generalized separable NMF and bi-orthogonal tri-factorization. Then, we proposed a convex optimization model to
tackle CoS-NMF, and developed a alternating fast gradient method to
solve the model. We compared the algorithms on
synthetic, document data sets and ORL facial image database. It is shown that
CoS-NMF model performs very well in co-clustering task, compared to the state-of-the-art methods. Some interesting interpretations of  CoS-NMF model for  applications on  document and image data sets are given and verified. 

Further work include to deepen our understanding of CoS
matrices which would allow us to design 
more efficient algorithms that provably recover optimal
decompositions in the presence of noise.


\bibliographystyle{abbrv}
\bibliography{nmfref,ref} 

\newpage
 \appendix
 \section{Appendix}
\subsection{Further Result for ORL facial database}
\begin{figure}[h]
 \centering
\includegraphics[width=1\textwidth]{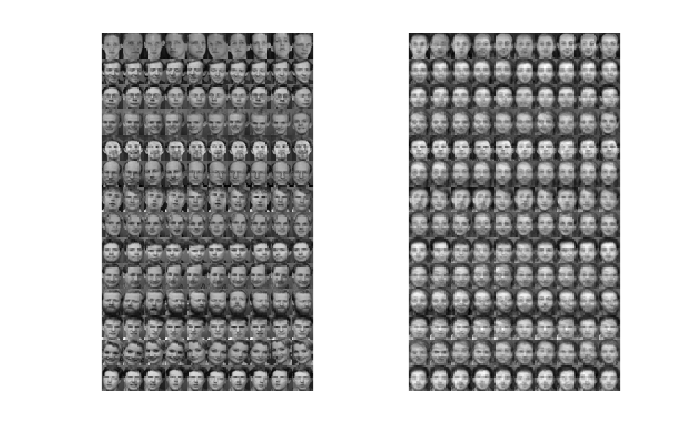}
\caption{Left:The 140 unselected facial images. Right: The reconstruction results. 
 The relative reconstruct approximation eror is $0.1847$.\label{unselect}}
\end{figure}

In ORL facial database, we notice that the facial images of \ref{key-subjects} are from 26 distinct subjects and assume that the other 14 subjects can be represented by these selected 40 facial images, denoted as $M(:,\mathcal{K}_2)$, where $\mathcal{K}_2$ is the index set of the selected facial images. The selected facial images set $M(:,\mathcal{K}_2)$ then can be regarded as feature. To verify our assumption, we reconstruct facial images of the  other unselected 14 subjects by $M(:,\mathcal{K}_2)$ and show the results in Fig.\ref{unselect}. It is interesting to see that the reconstruction facial images are similar to these 140 unselected  facial images, especially their hair colors, see the last two lines of facial images for example. The relative reconstruct approximation eror is $0.1847$. 

\subsection{Further Result for TDT30 document dataset}
In the following, we show the clustering results of words for TDT30 document dataset.

\begin{table*}[h]
  \centering
  \begin{tabular}{ c||c }
    \hline
    label&words\\
    \hline
   1 &  'case','lawyers', 'court', 'defense','law',    'lawyer', 'legal', 'judge', 'evidence', 'department',   'try', \\
   ~&'position', 'justice', 'tried', 'trial', 'order',   'hour', 'kaczynski', 'provide','begin', 'stand',\\
   ~&'suggested', 'claims','ruling', 'criminal', 'hearing',    'apparently', 'request','believed','understand'\\
 ~&'attempt', 'client', 'professor', 'prosecutor', 'appeal',    'ordered', 'rejected', 'decide'\\
    \hline
    2& 'game','left', 'jordan', 'night', 'point', 'lead',    'free',  'line', 'hit','center', 'gave',  'minutes', \\
    ~~& 'season','wanted',    'fourth',  'turned', 'series', 'helped', 'shot',   'michael', 'association',\\
    ~~& 'chicago',
    'account', 'bulls',  'knew', 'knows', 'giving'\\
    \hline
    3& 'lewinsky', 'told', 'starr', 'office', 'jury',   'monica', 'grand',  'relationship', 'job', 'according',\\
    ~& 'hours', 'call', 'testimony', 'tell',  'prosecutors',    'starrs',  'attorney', 'tripp', 'sources', 'friend',\\
   ~& 'brought', 'lewinskys', 'lie', 'investigators',   'ginsburg', 'offered', 'ken', 'worked', 'immunity',    \\
   ~& 'comment', 'truth', 'testify', 'agents', 'alleged',  'mother',    'conversations',  'whitewater',\\ 
   ~&'witness', 'linda',    'witnesses'\\
   \hline
   4&'nuclear', 'india', 'pakistan', 'tests', 'statement',    '\textbf{test}', 'arms', 'indias',  'indian',\\
   ~& 'bomb', 'response',  'sign', 'ban', 'missile',  'testing',  'range', 'declared',  'conducted'\\
   \hline
   5&'south', 'banks', 'korea', 'term', 'loans', 'debt',    'korean',  'due', 'banking'\\
   \hline
   6&'york', 'city', 'police', 'killed', 'street',   'authorities',  'town',  'simply', \\
   ~&'rules', 'italian',   'miles',  'mayor', 'streets',
    'officer', 'plane'\\
    \hline
    7& 'team', 'national', 'americans', 'led', 'russia',    'final',  'head', 'play', 'hard',\\
    ~& 'able', 'face', 'north', 'russian', 'teams',   'hockey',  'looking', 'martin', 'players', 'victory',\\
    ~& 'canada', 'round', 'allowed', 'leading', 'tour',    'ice', 'room', 'period', 'goal',\\
    ~& 'played', 'forward', 'coach', 'maybe', 'league',    'canadian', 'playing', 'damage', 'loss'\\
    \hline
    8&'sexual', 'federal', 'women', 'number','problem',    'found', 'pay', 'side', \\
    ~&'sex', 'drug', 'viagra','accused',  'makes',  'research',  'effect', \\
~&  'longer',    'ask',  'care',  'condition',  'mckinney',  'lives',    'cover'\\
\hline
9&'home',  'students', 'school', 'children', 'university',    'thought',\\
~& 'saw',  'fire',  'heard',  'student', 'started',    'stay'\\
\hline
10& 'government', 'economic', 'minister', 'foreign',    'economy',  'japan', 'financial',  'prime', 'saying',   \\ 
~& 'help', 'japanese', 'give', 'major', 'policy',  'past',   'problems',  'important',  'system', \\
~&  'hong',  'yen', 'central',  'exchange',    'deputy',  'finance',    'governments',  'bad',  'announced',  \\
~& 'kong',  'tax',   'cut', 'tokyo',  'budget',  'newspaper', 'confidence', 'turn',    'worlds', \\
~&  'means',  'key', 'development', 'role', 'domestic',  'ministry', 'bring', 'japans', 'measures',   \\
~& 'urged', 'especially',  'concern',  'yesterday',  'social', 'needs', 'ministers',  'forced',  'spending', \\
~& 'rest',    'package',  'asias', 'huge', 'fear', 'similar', 'concerned', \\
~& 'present',  'credit', 'steps', 'sector',
    'november'\\
\hline
11&'jones',  'mrs', 'working', 'involved', 'denied',    'paula',   'arkansas', 'lawsuit',  'provided'\\
\hline

   \end{tabular}  
 \end{table*}

\begin{table*}[h]
  \centering
  \begin{tabular}{ c||c }
    \hline
    label&words\\
    \hline
    12& 'companies','expected', 'including', 'group',   'business', 'big', 'company', 'times', 'workers',\\
~& 'making', 'added', 'find',  'seen',  'small',  'local',   'america', 'real', 'investment',\\
~& 'known', 'director', 'largest', 'union', 'sales',    'result',  'recently', 'technology', 'offer',\\
~&  'europe',  'share', 'jobs',  'cent', 'car', 'thai',    'parts',  'beginning',  'sell',\\
~&  'labor', 'services',  'single', 'products', 'build',   'heavy',  'success', 'costs',  'firm', 'noted'\\
\hline
    13& 'billion',  'million',  'months',  'prices',    'program',   'trade',  'oil',   'began',    'dlrs',\\
~&  'reported',  'food',  'current',  'european',   'increase',   'march',   'demand',   'dollars',   'aid',\\
~& 'total',  'islamic',  'buy',   'exports',   'cost',    'gas',    'imposed',    'algeria',    'production',\\
~&  'raised',   'poor',  'december',  'caused',  'daily',    'algerian',   'organization',   'export',    'amount'\\
\hline
14&  'put',  'capital',  'death',  'young', 'scheduled',    'woman',   'texas',  \\
~&  'person',  'cases', 'civil',  'penalty',  'florida',  'died', 'changed'\\
\hline
15&  'end', 'half',  'conference', 'super',  'chance',    'running',   'bowl',  'denver',  'green',  'pass'.\\
\hline
16& 'political',  'country',  'indonesia' ,    'leader', 'leaders',   'suharto',    'power',    'held',    'indonesian',\\
~& 'jakarta', 'countrys',  'family',  'forces',    'nation',    'future',    'soon',   'calls',    'opposition',\\
~& 'parliament',  'member', 'army', 'step',    'reform',  'indonesias',  'vice',  'reforms',    'leave',\\
~& 'rule',  'building',  'friends',  'cabinet',    'calling',   'rupiah',   'armed',    'planned',    'democracy',\\
~& 'hundreds',  'powerful', 'immediately', 'quoted',    'habibie',   'suhartos',   'hands',  'decades'\\
\hline
17&'spkr', 'news', 'today', 'voice', 'correspondent',    'look', 'voa',  'peter',\\
~& 'announcer',  'abc',  'jennings',  'camera',   'happen',   'tomorrow',  'jim',\\
~&  'phonetic',   'mark',   'sam',   'evening',    'voas',  'goes',   'tonight'\\
\hline
18&--\\
\hline 
19&'bank', 'meeting',  'israel',   'talks',    'peace',    'israeli',    'plan',   'albright',    'east',\\
~&  'process', 'netanyahu',  'pressure',   'control',    'move',   'middle',  'palestinian',   'meet',    'agreed',\\
~& 'west',  'met',  'london',   'effort',    'difficult',   'palestinians',   'hand', 'area',    'proposal',\\
~& 'hold',   'arafat',  'failed', 'negotiations' , 'authority',   'accept',    'land',    'idea',    'progress',\\
~&  'sides',  'agree', 'meetings',  'madeleine',    'areas',  'break',  'radio', 'summit', 'pro',    'blair'\\
\hline
20& 'tobacco',  'bill', 'money', 'industry',    'committee',  'campaign',  'senate',  'health',\\
~&  'anti', 'smoking', 'legislation', 'republicans',    'programs', 'settlement', 'reached', 'received',\\
~& 'debate', 'raise', 'tough', 'documents',  'proposed',  'cigarette', 'attorneys', 'fight',\\
~& 'june',  'related', 'interests',  'sen',  'democrats',  'cigarettes',  'age'\\
\hline
21&'visit',  'john', 'open',  'mass',  'cuba',   'pope',    'hope',    'change',\\
~& 'trip',  'cuban',  'history',  'community',    'paul',   'castro',   'arrived',  'thousands',\\
~& 'message', 'church',  'words',  'released',    'hopes',  'opportunity',  'freedom', 'society'.\\
\hline
22& 'united', 'states',  'american',   'officials',    'military',   'secretary',   'saddam',    'support',\\
~& 'war',  'gulf',  'force',  'action',  'hussein',    'region',  'attack',  'believe',\\
~&  'clear',  'air',  'situation',  'strike',   'diplomatic',    'continue',    'fact',\\
~& 'efforts', 'allow',  'arab',  'likely',  'based',    'british',   'kuwait',  'plans',\\
~&  'cohen',   'threat',   'troops',  'britain',    'chemical',    'mission',    'solution',    'relations',\\
~&  'stop',   'needed',  'strikes',   'william',    'ready',    'view',   'let',   'certainly',\\
~& 'spoke',    'french',    'speech',   'warned',    'comes',    'include',    'sense',    'ground',\\
~&  'pentagon',   'missiles',    'act',    'prepared',    'potential',    'wont',    'bombing',    'western',\\
~&  'aircraft',   'clearly', 'expressed',  'willing',    'allies',  'resolutions',   'ability', 'attacks',\\
~&  'suspected',  'persian',   'possibility',    'avoid',    'continues',   'diplomacy',   'prevent',    'saudi',\\
~&  'base',    'significant',  'send', 'carried',   'seek',  'carry'.\\
\hline

 \end{tabular}  
 \end{table*}
 
\begin{table*}[h]
  \centering
  \begin{tabular}{ c||c }
    \hline
    label&words\\
    \hline
    23&'crisis',  'international',  'asian',  'asia',    'countries',    'currency',    'imf',  'thailand',\\
~&  'monetary',  'malaysia',  'global',   'board',    'regional',   'singapore',   'economies',    'southeast',\\
~& 'currencies',  'philippines',  'worst',   'stability'.\\
\hline
24&percent', 'week', 'market',   'friday',    'high',    'month',    'stock',    'points',\\
~&  'report',   'chief',    'recent',    'weeks',    'fund',    'markets',    'dollar',    'growth',\\
~&  'earlier',   'interest',    'early',    'strong',    'close',    'investors',    'late',    'despite',\\
~&  'morning',    'low',    'price',    'rates',    'level',    'stocks',    'analysts',    'lost',\\
~& 'quarter',    'nearly',    'rate',    'fell',    'coming',    'large',    'lower',    'index',\\
~&  'higher',    'return',    'fall',    'funds',    'trading',    'remain',    'closed',    'average',\\
~& 'rose',   'main',    'earnings',   'impact',    'continued',   'inflation',    'january',    'remains',\\
~&  'ended',    'growing',    'expect',    'risk',    'rise',    'wall',    'showed',    'treasury',\\
~&   'corporate',    'biggest',    'previous',    'shares',    'performance',    'concerns',    'dow',    'hurt',\\
~&  'bond',    'april',    'turmoil',    'rally',    'drop',    'industrial',    'profits',    'annual',\\
~&  'securities',    'rising',    'analyst',    'dropped',    'february',    'signs',    'session'.\\  
\hline
25&'china',  'rights',    'chinese',    'human',    'live',    'beijing',    'chinas'.\\
\hline
26&'says', 'going','reporter',  'reports',    'headline',  'hes',  'show',  'cnn', 'kind',    'wants',\\
~& 'media',  'feel', 'mean', 'king', 'actually',   'happened',  'stories',  'weve',  'thank',  'believes',  'sort'.\\
\hline
27&'president',    'clinton',   'house',   'white',    'washington',    'public',    'called',    'part',    'official', 'asked',\\
~&   'story',  'clintons',  'trying',   'administration',  'investigation',  'issue',    'independent',   'need',   \\
~&  'presidents','question',   'counsel',  'decision',    'reporters',   'information',    'spokesman',    'press',   \\
~&  'senior',    'executive',  'issues', 'questions',  'television',   'material',    'matter',    'talk',    'private',    \\
~& 'service',    'follows',    'privilege',    'david', 'interview',  'intern',  'optional',   'chairman',    'charges',   \\
~&  'allegations',    'affair',    'republican',    'speaking',  'details',   'discuss',   'terms',  'kenneth',    'secret',      \\
~&  'reason',   'talking', 'personal',    'aides',     'democratic', 'staff',  'decided',  'quickly',   'scandal',   \\
~&   'weekend',    'seeking',    'attention',    'james',    'mike', 'refused',  'true',   'answer',  'spent',   'robert',  \\
~&  'appeared',  'claim',  'issued', 'wrong',   'consider',   'ways',  'letter',    'strategy',    'george',   \\
~&  'sought',    'investigating',    'protect', 'comments',    'focus',    'opinion',    'declined',    'inquiry',  \\
~&  'word',    'wife',    'affairs',    'accusations'.\\  
 \hline
  28& 'days',  'olympic',   'ago',   'top',    'games',    'won',    'place',    'set',    'nagano',    'olympics',\\
  ~& 'later',   'took',   'gold',   'course',    'win',    'run',    'medal',    'short',    'taking',   'race', 'winter', \\ 
  ~&    'record',    'sports',    'start',    'ahead',    'event',    'womens',    'cup',    'ski',    'events', 'slalom',\\
  ~&         'competition', 'figure',   'moment',    'training',    'italy',    'mens',    'seconds',    'downhill',  'finished',    \\
  ~&   'opening',    'conditions',   'athletes',    'considered',    'skating',    'site',    'champion',    'bit',   'germany',   \\
  ~&   'giant',    'felt',    'finally',    'snow', 'speed',   'finish',  'winning',    'cross',    'minute',    'sport',    'hill' \\
 \hline
 29& 'iraq',    'weapons',    'security',    'iraqi',    'nations',    'general',    'council',    'inspectors',    'baghdad',  \\
 ~&    'agreement', 'sanctions',  'annan',    'deal',    'sites',    'presidential',    'iraqs',    'inspections',    \\
 ~& 'special',    'butler',    'access', 'full',    'latest',    'biological',    'france',    'destruction',    'richard',    \\
~&'experts',    'kofi',    'iraqis'    'agency', 'inspection',   'commission',    'diplomats',  'resolution',     \\
~&  'ambassador',  'signed',   'inspector',    'cooperation',    'warning',   'palaces',    'richardson',    'standoff'.\\
\hline
30& 'congress',  'party',  'members',   'election',    'groups',   'front',  'violence',   'vote', 'form',  'hindu', \\
~& 'elections',   'majority',  'parties',     'politics',    'muslim',    'results',    'leadership',    'coalition',   'delhi'\\
\hline
 \end{tabular}  
 \end{table*}
\end{document}